\documentclass[10pt, a4paper, twocolumn, teaser, showabstract]{naverlabseurope}

\usepackage{multicol}
\usepackage{tikz,tkz-kiviat,pgfplots}
\pgfplotsset{compat=1.18}
\usepgfplotslibrary{groupplots}
\usepackage{xspace}
\usepackage{bm}
\usepackage{multirow} 
\usepackage{cleveref}
\usepackage{makecell}
\usepackage{subcaption}

\makeatletter
\def\onedot{\futurelet\@let@token\@onedot}
\def\@onedot{\ifx\@let@token.\else.\null\fi\xspace}

\def\eg{\emph{e.g}\onedot} 
\def\ie{\emph{i.e}\onedot} 
 
\def\etc{\emph{etc}\onedot}

\makeatother

\newcommand{\ours}{Multi-HMR 2\xspace}

\graphicspath{{fig/}}

\title{\ours: Multi-Person Camera-Centric Human \\ Detection, Mesh Recovery and Tracking}
\titlerunning{\ours}

\correspondingauthor{[guenole.fiche,fabien.baradel]@naverlabs.com}

\authors{
Guénolé Fiche
\authsep
Philippe Weinzaepfel
\authsep
Romain Brégier
\authsep
Fabien Baradel
}
\affiliations{NAVER LABS Europe}
\website{https://github.com/naver/multi-hmr2}
\websiteref{\href{https://github.com/naver/multi-hmr2}}

\teaserfig{\includegraphics[width=0.95\textwidth]{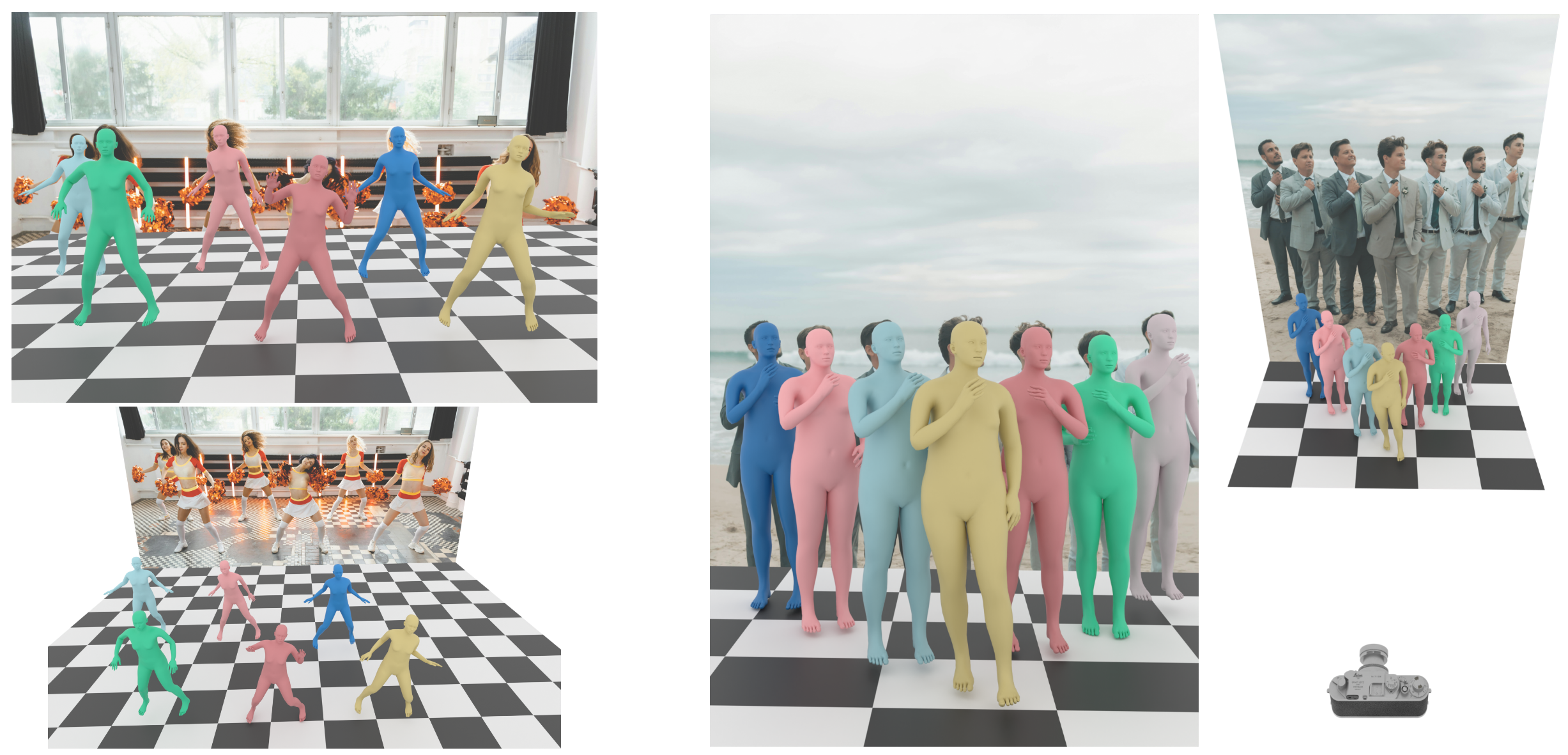}}

\teasercaption{\textbf{\ours} detects humans and recovers their 3D meshes, placed in the scene, along with camera parameters. 
  It also outputs per-human features that allow online tracking in videos, despite being trained only on still images.}

\begin{abstract}

Most advances in human mesh recovery (HMR) have focused on pelvis-centered recovery, overlooking metric 3D localization and detection accuracy in the camera coordinate system -- two key factors for real-world applications such as human–robot interaction and social scene understanding.
Current evaluation protocols often ignore these aspects, emphasizing per-person, root-centered recovery rather than camera-space perception.
As a result, existing approaches rely on fixed camera assumptions or handcrafted post-processing, limiting their robustness and practical deployment.
We introduce \textbf{\ours}, a simple yet robust DETR-based framework for \textbf{Mu}lti-person \textbf{C}amera-centric \textbf{H}uman detection, m\textbf{e}sh \textbf{R}ecovery, and tracking.
\ours predicts a scene-consistent camera together with human meshes, enabling metric 3D localization without ground-truth intrinsics.
Moreover, by distilling image-based memory features from SAM2, \ours extends to tracking, achieving consistent identity association without video supervision.
Despite its conceptual simplicity -- no handcrafted components, no video input, and no ground-truth cameras -- \ours achieves state-of-the-art pelvis-centered performance while substantially improving detection accuracy and metric 3D localization.

\end{abstract}

\begin{document}

\maketitle

\section{Introduction}
\label{sec:intro}

Perceiving humans from images is a core computer vision problem, with applications in diverse fields such as robotics, virtual reality, or automatic gesture analysis. Among human perception tasks, multi-person human mesh recovery (HMR) aims to detect, localize, and estimate the 3D mesh of every human in a scene. Although useful for understanding human interactions and for robotic navigation applications, this task is particularly challenging as one needs to deal with depth ambiguity, scene and inter-person occlusions, and computational cost constraints when recovering multiple humans simultaneously. Earlier works in multi-person HMR~\cite{choi2022learning, qiu2022dynamic, jiang2020coherent, zanfir2018monocular} followed a two-step procedure consisting of: (1) detecting every human in the scene, typically with a pretrained object detection model~\cite{detectron2, maskrcnn, ssd, yolo}; (2) estimating a mesh independently for every human by cropping the image around each person~\cite{hmr, humans4d, camerahmr, pymaf, smplerx}.
While enabling the use of powerful foundation models for object detection and single-person HMR, this paradigm presents several drawbacks: the inference cost grows with the number of individuals in the image, interactions between humans rely exclusively on global image features, and complex procedures are designed to localize humans in the 3D scene.
To address these limitations, recent approaches propose to jointly recover all human meshes given an image using one-stage detectors.

Following the common practices in object detection, one-stage approaches to multi-person HMR can be divided into 2 categories: (1) In the spirit of CenterNet~\cite{centernet}, some approaches rely on local image features to perform detection and feed the relevant ones directly into the decoder,
(2) DETR-like models~\cite{detr} have a fixed set of queries and need to perform query to target matching at training time and query selection at test time. Despite sometimes showing slower training convergence, DETR approaches typically better capture the global scene context as queries do not depend on local image features. Additionally, CenterNet-like approaches such as Multi-HMR~\cite{multihmr} cannot handle images with close interactions when two humans are located in the same image patch, as the query would be the same. AiOS~\cite{aios} and SAT-HMR~\cite{sathmr} are DETR-based approaches.
However, these methods make very strong assumptions about the camera model, with a fixed focal length of 5000~mm in~\cite{aios} and a fixed field of view (FOV) of $\frac{\pi}{3}$ for~\cite{sathmr}. This assumption leads to large errors in the estimate of 3D locations in the camera frame. Additionally, multi-person HMR methods~\cite{aios, sathmr, multihmr} use parametric body models that can only represent adult bodies~\cite{smplx, smpl}: this leads to an increased 3D localization error in the presence of children, which are predicted as adults at a long distance from the camera to compensate for the scaling error.

\begin{figure*}
    \centering
    \includegraphics[width=0.7\linewidth]{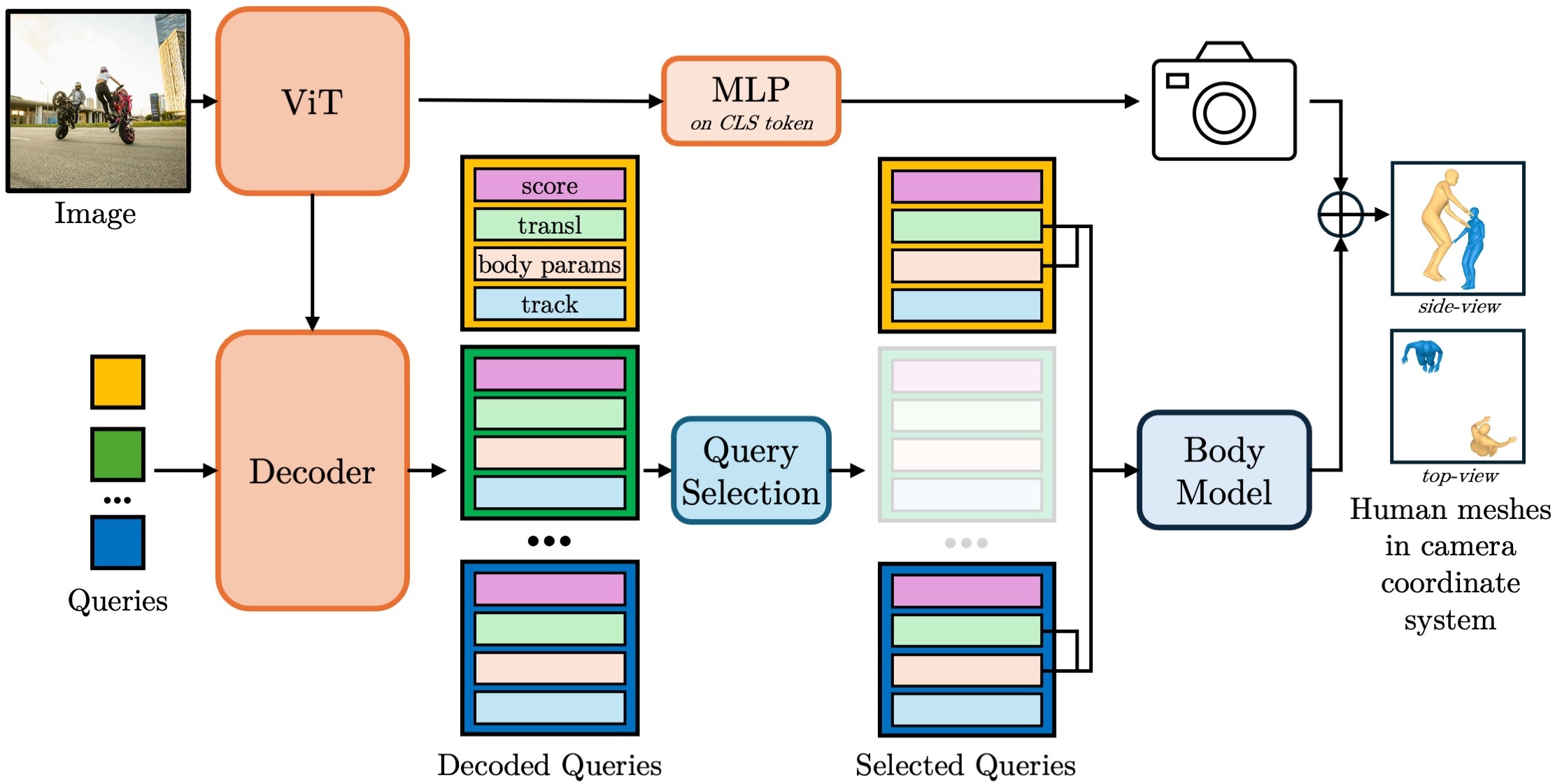} \\
    \caption{\textbf{Overview of \ours}. \ours follows a DETR-based architecture composed of an encoder that extracts image tokens and predicts camera parameters, and a decoder that cross-attends learned queries with image tokens before outputting score, 3D location, body parameters and tracking features for each query. A subset of queries is selected with an Hungarian matching at training or by thresholding the score at inference, and is fed to a body parametric model to obtain the final meshes.}
    \label{fig:overview}
\end{figure*}

In this work, we introduce \ours, a multi-person HMR approach leveraging the Anny~\cite{anny} parametric body model. Anny is open source and covers the human diversity — across ages (from infants to elders), body types, and proportions — with a unified model featuring interpretable body shape parameters. For detecting humans we rely on a DETR framework (see Figure~\ref{fig:overview}): this makes \ours a strong human detector, robust to heavy occlusions and to humans in close interaction. One of the limitations of DETR-based models is that training can be computationally expensive, depending on the modalities used to associate the predictions with the targets in the Hungarian matching algorithm.
In contrast with~\cite{sathmr, aios}, we propose to match predictions with targets based solely on the 2D location of humans, avoiding costly forward and backward passes through the parametric body model for all queries.
We show that this does not affect the HMR performance while allowing us to use additional sources of training data without ground-truth meshes and significantly reducing the compute needs for training. \ours is trained in around a week on a single GPU.

By contrast to~\cite{sathmr, aios}, we additionally propose to estimate the camera intrinsic parameters. This allows us to localize humans much more accurately in the camera space, both in terms of absolute distance to the camera and relative distance between humans. Combined with the ability to model children, \ours presents unprecedented capability in metric-scale recovery of human-centered scenes from a single image. 

In addition to 3D properties, we propose to regress features distilled from the SAM2~\cite{sam2} memory encoder for each human, that can be used to track humans by associating detections across frames. This makes \ours the first approach for human tracking that can be trained without any video or mesh sequence.

We evaluate \ours on in-the-wild HMR benchmarks~\cite{3dpw, emdb}, humans in close interaction~\cite{hi4d, harmony4d}, reprojected keypoints~\cite{coco}, and on the tracking task~\cite{posetrack21}. We show that despite its simplicity, \ours achieves results superior or comparable to the SOTA multi-person HMR methods.

In summary, we make the following contributions:
\begin{itemize}[leftmargin=*]
    \item We introduce \ours, a multi-person HMR method based on the inclusive Anny body model.
    \item \ours is based on the DETR framework, making it robust to overlapping humans. Our novel matching strategy based on 2D location 
    makes \ours fast to train and trainable on images without 3D annotations.
    \item \ours can use the ground-truth camera parameters or predict them, allowing highly-accurate human localization in the scene.
    \item We distill SAM2 features, allowing \ours to track humans by associating detections across frames without any training on video data.
\end{itemize}
\section{Related work}
\label{sec:relatedwork}

\paragraph{Multi-person HMR.} 
Human mesh recovery, introduced in~\cite{hmr}, is the task of regressing human meshes given an image. In this work we focus on parametric methods, that predict the parameters of a human body model~\cite{smpl, smplx, ghum, anny}. While most methods focus on the single-person scenario and assume that their input is an image tightly cropped around the human of interest, more and more works shift to the multi-person HMR task. Multi-person HMR methods aim to detect and localize humans in the 3D scene in addition to regressing human meshes. Earlier approaches were multi-stage~\cite{choi2022learning, qiu2022dynamic, jiang2020coherent, zanfir2018monocular} and relied on off-the-shelf human detectors~\cite{detectron2, maskrcnn, ssd, yolo} and single-person HMR methods~\cite{hmr, humans4d, camerahmr, pymaf, smplerx}. However, this multi-stage paradigm lead to an inference time linearly growing with the number of subjects in the image, and ignored the spatial relationships and interactions between humans. Following the seminal work of ROMP~\cite{romp}, some single-stage approaches were proposed~\cite{bev, psvt, multihmr}. These approaches are inspired by the CenterNet~\cite{centernet} paradigm: they first detect humans and then use their local image features as queries to predict meshes. Recently, PromptHMR~\cite{prompthmr} united the CenterNet and multi-stage paradigms by enriching Multi-HMR with ground-truth or off-the-shelf detections, segmentation, and textual information. However, this kind of approaches typically ignores the global context, as queries contain most of the necessary information to predict human meshes. Additionally, they are not optimal for handling overlapping humans. For these reasons, recent work explores DETR-like approaches for multi-person HMR~\cite{aios,sathmr}.

In this work, we propose a new DETR-based approach to multi-person HMR. The DETR framework allows end-to-end training, better takes into account the global image context, and provides an efficient and elegant way to deal with detections, especially in occlusion scenarios. 

\paragraph{DETR for human perception.} 
The DETR architecture~\cite{detr} was originally proposed for object detection. It consists of feeding a set of learned queries to the decoder. Each query will result in a prediction, with a given confidence value. During training, predictions are matched with the ground truth using a Hungarian algorithm. At inference, queries are selected using the predicted confidence value. Many methods built upon DETR~\cite{groupdetr, deim, dabdetr} improved its convergence, and first applications to human perception were made in 2D pose estimation~\cite{grouppose, shi2022end, yang2023explicit}. Starting from ED-Pose~\cite{yang2023explicit}, AiOS~\cite{aios} was the first DETR-based approach to multi-person HMR. AiOS first localizes subjects with a human token and introduces joint-related tokens to predict expressive human meshes~\cite{smplx}. However, although being one-stage, AiOS relies on crops to predict fine-grained details such as hands and facial expression, and it assumes a focal length of 5000~mm, making the global localization prediction inaccurate. SAT-HMR~\cite{sathmr} extends Multi-HMR~\cite{multihmr} to the DETR framework with scale-adaptive image features. SAT-HMR achieves performance comparable with~\cite{multihmr, aios} in pelvis-centered metrics with a minimal inference cost, however, its predictions in the camera coordinates system lack accuracy as it relies on cameras with a fixed FOV.

\ours focuses on improving existing multi-person HMR methods in terms of 3D localization in the camera frame. For that, we propose an adaptive model that can leverage ground-truth camera parameters or predict the FOV. We show that \ours achieves SOTA performance in pelvis-centered metrics while significantly improving scene-centered metrics.

\paragraph{Human tracking.}
Multi-person HMR can be extended to videos, which requires tracking humans across frames to leverage the temporal context. Most approaches to human tracking work in a tracking-by-detection setting, which means that they detect objects in every frame independently and then associate them. Earlier approaches were based on 2D locations and keypoint features~\cite{detectNtrack, xiao2018simple}. Incorporating HMR predictions into the tracking pipeline significantly improves accuracy~\cite{hmar, phalp}, and~\cite{humans4d} show that stronger detections and human mesh recovery estimates lead to even better tracking. Recently, CoMotion~\cite{comotion} proposed a tracking-by-attention method that detects new objects and updates existing tracks jointly, and~\cite{motionqueries} proposed a DETR-based approach that considers the predictions of a given query across time as a motion prediction.

While prior works all rely on video or human motion data for training, we propose to distill SAM2 features, that can be used for tracking by associating detections across frames while being trained only on still images.
\section{Method}
\label{sec:method}

\begin{figure*}[t]
    \centering
    \includegraphics[width=0.8\linewidth]{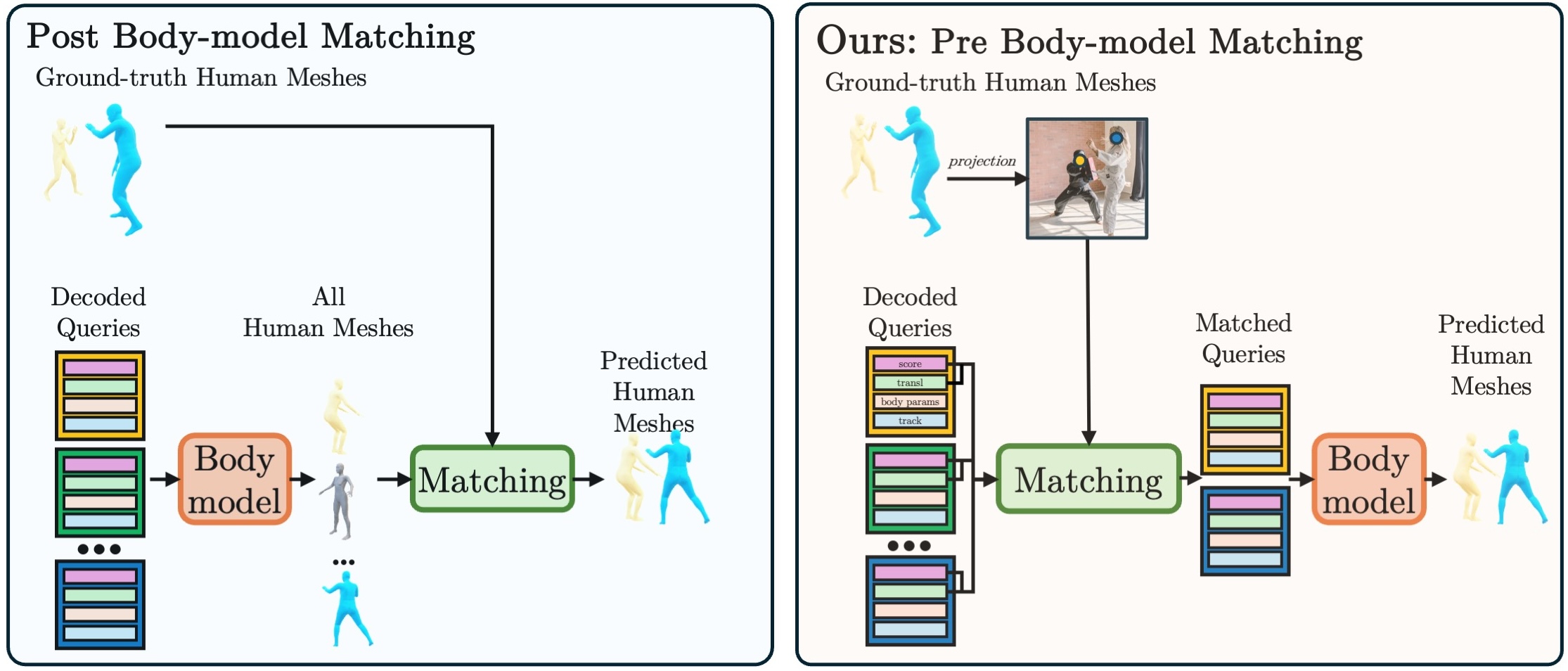} \\[-0.3cm]
    \caption{\textbf{Matching queries and ground-truth humans during training.} Prior works~\cite{aios,sathmr} fed all queries to the body model to perform matching using all joints (left). In contrast, we perform the matching using the location only (right): only the matched queries require a forward pass through the body model, significantly reducing computational cost and accelerating training.}
    \label{fig:matching}
\end{figure*}

\paragraph{Overview.}
Figure~\ref{fig:overview} shows an overview of the \ours's architecture, that given a single RGB image, detects humans in the scene and regress their meshes in the camera coordinate system.
The model is based on the DETR~\cite{detr} paradigm for object detection: the image is encoded with a ViT.
A decoder based on cross-attention blocks with respect to the encoded tokens then processes a set of learned queries.
These decoded queries are then fed to different MLP heads: a first one predicts a detection score, a second one predicts global location, a third one predicts the parameters of a parametric human body model, and a last one predicts a feature that can be used for tracking. We additionally regress the camera intrinsics from the \emph{cls} token of the encoder.

DETR-like detectors have the advantage of elegantly dealing with overlapping humans, in contrast to CenterNet-like approaches~\cite{bev,multihmr} or two-stage methods~\cite{humans4d}. While they might be slow to train, in particular if all queries are forwarded to the body model during training, we propose to perform the matching considering the body location only, avoiding this problem
(see Figure~\ref{fig:matching}). By using only the 2D localization in pixels, we are also able to train using datasets annotated with 2D keypoints alone~\cite{coco}.

As parametric body model, we rely on the recent Anny~\cite{anny}, for its strong ability to model human diversity (in terms of ages covering babies to elders, genders, weight,
\etc).
As most other parametric models~\cite{smplx,ghum}, Anny encodes human meshes, including face and hands, using only a few  bone 3D orientations, namely 163 and a set of 6 interpretable body shape parameters.

As feature relevant for tracking, we propose to distill the flattened feature maps of the SAM2~\cite{sam2} image and memory encoder: this can be trained from still image supervision, \ie, without any video annotated with human meshes.
This stands in contrast to classical feature obtained from appearance cues~\cite{detectNtrack,phalp}.

We now detail the components of our model.

\paragraph{Image Encoder.}
The input image is processed with a ViT-Large encoder~\cite{ViT}: the image is split into patches, which are fed to a series of transformer blocks~\cite{transformer}. A linear layer maps the obtained image features to the dimension of the decoder.
While most HMR images process only square images, by padding the borders, we train our model with various aspect ratios.
We crop the images during training to randomly chosen aspect ratio while the largest dimension is set to 768 pixels. The crops are centered to preserve the centering of the principal point. At inference, the image is simply resized such that the largest dimension is 768 and the smallest side a multiple of the patch size.

\paragraph{Camera intrinsics prediction.}
We use a Multi-Layer Perceptron (MLP) head to predict the camera intrinsic parameters from the \emph{cls} token of the ViT encoder. More precisely, we regress the vertical FOV, assuming similar focal length along horizontal and vertical axis, and a principal point at the center of the image.

\paragraph{Decoder.}
The \ours's decoder follows a DETR-like architecture. A set of 100 learned human queries is decoded through 8 decoder blocks, each composed of a multi-head self-attention layer between queries, a multi-head cross-attention layer with image tokens, and an MLP. The decoded queries will then be given as input to multiple prediction heads, each responsible for the prediction of a given modality, as detailed below.

\paragraph{Detection score.} Our model predicts a detection score for each decoded query. This score will be used in the matching process during training, while at inference time, we only keep queries with a score above a given detection threshold.

\paragraph{Location prediction.}
Regressing the 3D location of humans in the camera space is particularly challenging, especially when focal length is not fixed. We decompose the global location estimation into 2 simpler problems. First, we estimate the 2D location of a human in pixel space. Following the monocular depth literature~\cite{ranftl2020towards, mertan2022single, facil2019cam} and prior works in multi-person HMR~\cite{multihmr, prompthmr} we then predict the nearness, that corresponds to the depth in log-space. We can then recover the 3D location by inverse projection using our predicted depth and 2D location, and estimated camera intrinsics.

\begin{figure*}[t]
    \centering
    \includegraphics[width=\linewidth]{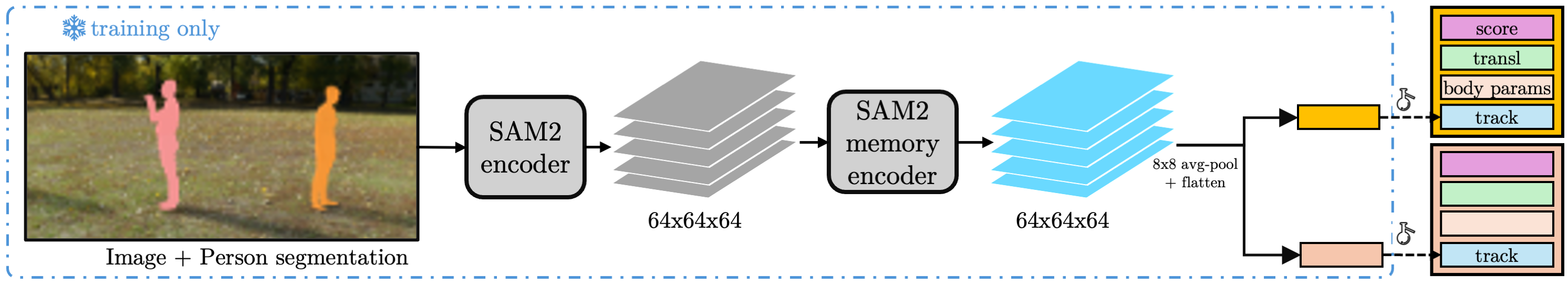} \\[-0.2cm]
    \caption{\textbf{Target tracking feature.}
    We obtain a target feature for each person using SAM2.
    The image and ground-truth mask are fed into SAM2, and the average-pooled memory encoder output is flattened into a feature vector.
    }
    \label{fig:trackfeat}
\end{figure*}

\paragraph{Matching queries and ground-truth human meshes during training.}
Since our model is based on the DETR framework, predictions need to be matched with the ground-truth meshes during training. Unlike prior DETR-like HMR methods~\cite{aios, sathmr}, we propose to do the Hungarian matching based solely on the predicted confidence and 2D location, before predicting meshes. Let us denote $\hat{y}{=}\{\hat{l}_i, \hat{c}_i\}_{i=1}^N$ the $N$ predicted locations in pixels and confidence values, and $y{=}\{l_i\}_{i=1}^M$ the $M$ target locations.
Let ${x_{ij}}$ be the binary variables
indicating if the prediction $i$ matches the target $j$. We solve the following problem:
\begin{gather}
    \underset{\{x_{ij}\}}{min} \sum_{i=1}^N \sum_{j=1}^M x_{ij} \left(\lVert \hat{l}_i - l_j\rVert_1 + FC(\hat{c}_i)\right)
\\
\text{s.t.}~~~~\sum_{j=1}^M x_{ij}{\leq}1~~~\forall i{=}1...N, ~~~~~~ \sum_{i=1}^N x_{ij}{=}1 ~~~\forall j{=}1...M, \nonumber
\end{gather}
with $FC(\hat{c}_i) = -\alpha(1-\hat{c}_i)^\gamma\log(\hat{c}_i)$ a focal loss~\cite{focalloss} with $\alpha{=}0.25$ and $\gamma{=}2$.

This 2D-based matching strategy offers two key advantages. First, it greatly reduces the computational cost, since only matched queries are passed through the differentiable parametric body model, while unmatched queries are discarded early as shown in Figure~\ref{fig:matching}. A forward-backward pass takes 0.35s with our matching strategy vs. 0.48s with the standard one. It also uses much less memory (7GB vs. 27GB), enabling larger batch sizes and further speed gains.
Second, it enables training with images annotated solely with 2D keypoints or bounding boxes, as no 3D mesh supervision is required for the matching step.
In Table~\ref{tab:ablations} of Section~\ref{sec:experiments}, we demonstrate through an ablation study that this strategy maintains comparable accuracy to 3D-based matching while significantly improving training efficiency and flexibility across heterogeneous supervision sources.

\paragraph{HMR prediction.}
For each matched query, we predict the Anny shape parameters and the joint angles in the form of 6D rotations~\cite{6drot} that correspond to relative angles along the kinematic tree. The Anny body model then outputs the mesh vertices and joint locations in a differentiable fashion.

\paragraph{Regressing track features.} 
We aim to also obtain an appearance feature per human that would allow to associate detections across frames in videos at inference, thus allowing to track people.
One possible solution for that is to build an explicit texture map~\cite{phalp,humans4d,hmar} but this might suffer from inaccurate mesh estimation or varying illuminations. An alternative is to rely on contrastive learning, inspired by the person re-identification literature~\cite{luo2019bagoftricks,almazan2018re}, but this would require pairs of images; or distillation of a strong already trained model. However, these models are known to suffer from dataset biases~\cite{zheng2016person,zhong2018camera}. In this work, we propose to distill the knowledge from the SAM2 foundation model for image and video segmentation~\cite{sam2} (see Figure~\ref{fig:trackfeat}).

SAM2 processes each video frame independently, but it maintains a memory bank that stores short-term spatio-temporal information from recent frames. For a new frame, the model first encodes the image into tokens. These tokens then interact, via cross-attention, with the memory bank, which holds features from previous frames. A mask decoder uses the resulting features to generate the segmentation mask for the current frame.
After producing the mask, SAM2 encodes both the mask and the memory-enhanced features from the current frame to create new feature representations.
These features are added to the memory bank and correspond to a $64 \times 64$ spatial grid of features, each of dimension $64$. The memory bank contains the set of features from the last $7$ frames at maximum.

In our case, given a training image with ground-truth human segmentations, we build a tracking feature (as shown in Figure~\ref{fig:trackfeat}) by (a) encoding the image with the SAM2 image encoder thus obtaining an image feature map, (b) feeding that image feature map with the ground-truth segmentation to the memory encoder to obtain a memory feature map, (c) compressing this $64 \times 64 \times 64$ representation into a $4096$-dimensional vector using an $8 \times 8$ average pooling and a flattening operation. This feature can be computed for every human in training images as long as ground-truth segmentations are available, either with annotations~\cite{coco} or thanks to the synthetic nature of many datasets~\cite{bedlam,anny}.

\paragraph{Training losses.}
Training losses include a cross-entropy for the predicted confidence, and $L_1$ losses for the predicted 2D locations, depths, pose parameters, shape parameters, 3D vertices and joint positions centered on the pelvis, re-projected 2D joint coordinates using the ground-truth or predicted camera, and regressed track features. Following~\cite{camerahmr}, we use an asymmetric loss for the predicted FOV, which penalizes overestimated values more.

In order to improve the stability of training, we additionally use a stable loss for 2D joints based on the reprojection error on a sphere. The inverse projection operator $\Pi^{-1}_{\bm{K}}$ that maps 2D camera coordinates to 3D directions on the half-sphere in front of the camera is continuous and differentiable, contrary to $\Pi_{\bm{K}}$. We therefore minimize the angular error between the predicted 3D directions and target rays:
\begin{equation}
    \mathcal{L}_{\text{2D angle}} = \cfrac{1}{s_{\bm{K}}} \sum_i \angle(\bm{x}_i, \Pi_{\bm{K}}^{-1}(\hat{\bm{u}}_i))
\end{equation}
where $s_{\bm{K}}$ is a scaling factor defined as the camera diagonal field of view angle. It is stable except for points located at the camera center, where the angle is undefined; we thus ignore points within a $0.1$m radius around the camera center when computing this loss.

\paragraph{Dealing with heterogeneous training data.}
Our training relies on a mixture of synthetic data, pseudo-ground-truth (pseudo-GT) meshes on real images, and real images annotated with only 2D keypoints. We adapt the supervision strategy for each data type based on its reliability and annotation quality.

For \textit{synthetic data}, which provides accurate ground-truth meshes and cameras, we supervise all prediction heads: camera parameters, detection confidence, 2D locations, depth, body pose, shape, vertices and their 2D reprojections.
For \textit{pseudo-GT real images}, we supervise only the 3D body pose and shape, but not the detection confidence, since the annotations are sparse and do not cover all people in the scene. This prevents penalizing valid detections that lack annotations.
Finally, for \textit{real images with 2D keypoint annotations only}, we supervise the detection heads and the 2D re-projection of certain joints (\ie MS-COCO). This encourages alignment between projected 3D keypoints and annotated 2D joints, enabling training even without mesh supervision.

This adaptive supervision strategy allows \ours to effectively combine heterogeneous datasets with varying annotation quality, ensuring robustness and generalization across both synthetic and real data.

\paragraph{Training details.}
The encoder weights are initialized with DINOv3~\cite{dinov3}, while other parts of the model are trained from scratch.
We use the Adam optimizer~\cite{adam} for training.
As training data, we use the Anny-One~\cite{anny} and BEDLAM~\cite{bedlam} synthetic datasets, as well as real images from MSCOCO~\cite{coco}, MPII~\cite{mpii} and AIC~\cite{aic} with the pseudo-ground truth meshes from CameraHMR~\cite{camerahmr}. For all datasets except Anny-One, that is already in the Anny format, we fit the Anny model to existing SMPL~\cite{smpl} annotations. This model pretrained exclusively with 3D annotations is denoted \textbf{\ours.a} in the experiments section.

One of the weaknesses of the pseudo-ground truth meshes is that they are sparse, which can lead to poor detection performance on challenging in-the-wild images. 
We experiment using additional 2D large-scale pseudo-ground truth on MSCOCO~\cite{coco} and Openimages~\cite{openimages}, obtained using a strong detector~\cite{detectron2} and 2D pose estimation method~\cite{vitpose}. However, we observe that naively mixing 2D and 3D supervision alters the pose prior learned during 3D training. In particular, when lower-body joints are occluded, the model frequently predicts sitting poses. We observe a similar behavior in SAT-HMR~\cite{sathmr}, suggesting that this issue is not specific to the architecture but rather is a consequence from sparse 2D supervision. To mitigate this undesired effect, we propose to regularize the finetuning stage towards the predictions of the \ours.a model. Specifically, we add consistency $L_2$ losses on pose parameters and 3D joints, together with a reprojection loss on COCO joints when 3D annotations are available.
The finetuned model is denoted as \textbf{\ours.b} in the experiments section.

For the target tracking features obtain from SAM2~\cite{sam2}, we use its Hiera-L version~\cite{hiera}.

\paragraph{Inference on images.}
At inference, the image is padded to the closest aspect ratio used for training, and resized so that its largest dimension is 768. Instead of doing Hungarian matching on the 2D location, we keep the queries with a confidence value greater than 0.4. Optionally, a non-maximum-suppression is done based on the 3D location.

\paragraph{Tracking in videos.} 
In the case of videos, we process the frames one by one and detect humans and their meshes with our model. Following methods based on per-frame appearance features, \eg~\cite{phalp}, we build a similarity matrix between human detections in the current frames and active human tracks from previous frames and
finally use the Hungarian matching algorithm to associate an ID to each detection in the current frame. We now give more details about the similarity score we use.

We first build a similarity matrix between all new detected humans and all active human tracks based on the predicted features distilled from SAM2. To this end, we keep a bank of features seen in the past 50 frames, together with their assigned IDs. For each new feature, we use K-Nearest-Neighbors (KNN) search to retrieve the most similar ($K{=}35$) in the bank according to the L1 distance, apply a softmax function with a temperature parameter of $t{=}15$, and aggregate the score for each human ID.

Second, for each human ID, we keep track of the last 8 positions and orientations of the pelvis keypoint in 3D space, and we use a ridge regression, to predict the expected position and orientation of the pelvis in the current frame. Let $\hat{p}_j,\hat{n}_j,\hat{\theta}_j$ be the predicted pixel, nearness and orientation respectively for a given human ID $j$. Given a new human detection $i$ and its pelvis pixel $p_i$, nearness $n_i$ and orientation $\theta_i$, we build a similarity matrix between each new detection and every human ID already assigned using the following formula: $e^{-\Vert p_i - \hat{p}_j \Vert /t_p} e^{-\Vert n_i - \hat{n}_j \Vert /t_n} e^{-\Vert \theta_i - \hat{\theta}_j \Vert / t_{\theta}}$ with $t_p{=}1$, $t_n{=}3$ and $t_{\theta}{=}20$.

The two similarity matrices are combined to obtain a score per association using a weighted sum with weights of 1 (resp. 8) for the feature (resp. pelvis) similarity. Once the similarity matrix is computed, we use an Hungarian matching algorithm to make the association. We stop a track when it had no association in the last $50$ frames. We start
new tracks when some new detections have no association with a similarity below $0.69$.

\section{Experiments}
\label{sec:experiments}

We compare \ours to the state of the art in Section~\ref{subsec:SOTA-comparison} and then ablate its main components in Section~\ref{subsec:ablations}.

\paragraph{Evaluation benchmarks.} Following recent work in HMR, we evaluate the in-the-wild HMR performance of \ours on the  3DPW~\cite{3dpw} and EMDB~\cite{emdb} benchmarks. We also use the Hi4D~\cite{hi4d}, Harmony4D~\cite{harmony4d}, and CMU-Panoptic~\cite{joo2015panoptic} test sets, which focus on people in close interaction, to evaluate the detections under occlusion and the capacity of the model to capture the relative position of humans in the scene. We evaluate the reprojection on the MSCOCO~\cite{coco} dataset, and for the tracking task, we test on the PoseTrack21~\cite{posetrack21} validation set.

\begin{table*}[t]
\centering
\caption{\textbf{Evaluation on HMR benchmarks} (3DPW and EMDB).} 
\vspace{-0.3cm}
\small
\begin{tabular}{lc|c@{~~}c@{~~}c@{~~}c|c@{~~}c@{~~}c@{~~}c}
\toprule
\multirow{2}{*}{\textbf{Method}} &
\multirow{2}{*}{\textbf{Camera}} &
\multicolumn{4}{c|}{\textbf{3DPW}} &
\multicolumn{4}{c}{\textbf{EMDB}} \\
 & & {\scriptsize PA-MPJPE $\downarrow$} & {\scriptsize MPJPE $\downarrow$} & {\scriptsize PVE $\downarrow$} & {\scriptsize TA-PVE $\downarrow$}
 & {\scriptsize PA-MPJPE $\downarrow$} & {\scriptsize MPJPE $\downarrow$} & {\scriptsize PVE $\downarrow$} & {\scriptsize Abs-PVE $\downarrow$} \\
\midrule
AiOS~\cite{aios} & Fixed & 45.0 & 68.8 & 90.9 & 774.3
 & 63.3 & ~~90.6 & 108.1 & 11962.1 \\
SAT-HMR~\cite{sathmr} & Fixed & 52.7 & 81.0 & 94.5 & 174.5 
 & 71.0 & 112.9 & 126.7 & ~~1093.0 \\
\textbf{\ours.a (ours)} & Predicted & \textbf{40.5} & \underline{68.2} & \underline{80.1} & \underline{170.1}
 & \textbf{48.3} & ~\textbf{68.5} & ~~\textbf{78.7} & ~~~~\textbf{273.6} \\
\textbf{\ours.b (ours)} & Predicted & \underline{41.8} & \textbf{65.2} & \textbf{78.1} & \textbf{161.9}
 & \underline{52.5} & ~\underline{71.0} & ~~\underline{83.6} & ~~~\underline{374.8} \\ \midrule
Multi-HMR~\cite{multihmr} & GT & 46.9 & 69.5 & 88.8 & 187.2
 & \underline{48.5} & ~~73.7 & ~~87.1 & ~~~~168.1 \\
\textbf{\ours.a (ours)} & GT & \textbf{40.2} & \underline{67.7} & \underline{79.9} & \underline{149.8}
 & \textbf{48.3} & ~~\textbf{68.5} & ~~\textbf{78.7} & ~~~~\textbf{129.4} \\
\textbf{\ours.b (ours)} & GT & \underline{41.6} & \textbf{64.5} & \textbf{77.7} & \textbf{145.6} 
 & 52.5 & ~~\underline{71.0} & ~~\underline{83.6} & ~~~~\underline{131.7} \\
\bottomrule
\end{tabular}
\normalsize
\label{tab:3dpw-emdb}
\vspace{-0.1cm}
\end{table*}

\begin{table*}[t]
\centering
\caption{\textbf{Evaluation on interaction HMR benchmarks} (Hi4D and Harmony4D).}
\vspace{-0.3cm}
\small
\begin{tabular}{lc|c@{~~}c@{~~}c@{~~}c|c@{~~}c@{~~}c@{~~}c}
\toprule
\multirow{2}{*}{\textbf{Method}} &
\multirow{2}{*}{\textbf{Camera}} &
\multicolumn{4}{c|}{\textbf{Hi4D}} &
\multicolumn{4}{c}{\textbf{Harmony4D}} \\
 & & {\scriptsize MPJPE $\downarrow$} & {\scriptsize Pair-PA-MPJPE $\downarrow$} & {\scriptsize TA-PVE $\downarrow$} & {\scriptsize F1-score $\uparrow$}
 & {\scriptsize MPJPE $\downarrow$} & {\scriptsize Pair-PA-MPJPE $\downarrow$} & {\scriptsize TA-PVE $\downarrow$} & {\scriptsize Recall $\uparrow$} \\
\midrule
AiOS~\cite{aios} & Fixed & 72.9 & 239.0 & 694.4 & ~~98 
 & \textbf{130.5} & 304.8 & 3454.8 & 91 \\
SAT-HMR~\cite{sathmr} & Fixed & 88.2 & ~~85.5 & 132.4 & \textbf{100} 
 & 140.3 & 144.0 & ~~676.3 & \textbf{97} \\
\textbf{\ours.a (ours)} & Predicted & \textbf{61.7} & ~~\textbf{79.5} & \textbf{114.9} & \textbf{100} 
 & \underline{131.8} & ~~\textbf{99.8} & ~~\textbf{231.2} & \textbf{97} \\
\textbf{\ours.b (ours)} & Predicted & \underline{65.9} & ~~\underline{82.9} & \underline{123.5} & ~~\underline{99}  
 & 132.4 & ~~\underline{105.4} & ~~\underline{382.1} & \underline{96} \\ \midrule
Multi-HMR~\cite{multihmr} & GT & 69.2 & ~~82.2 & \underline{116.2} & ~~98 
 & 146.0 & 140.8 & ~~388.1 & \underline{93} \\
\textbf{\ours.a (ours)} & GT & \textbf{61.7} & ~~\textbf{78.3} & \textbf{113.1} & \textbf{100} 
 & \textbf{128.4} & ~~\textbf{96.2} & ~~\textbf{233.6} & \textbf{97} \\
\textbf{\ours.b (ours)} & GT & \underline{65.9} & ~~\underline{80.5} & 119.5 & ~~\underline{99}
 & \underline{128.7} & ~~\underline{96.7} & ~~\underline{274.6} & \textbf{97} \\
\bottomrule
\end{tabular}
\normalsize
\label{tab:hi4d-harmony4d}
\vspace{-0.2cm}
\end{table*}

\paragraph{Metrics.} We report standard HMR metrics in~mm: the per-vertex error (PVE) measures the Euclidean distance between the vertices of the predicted mesh and the ground truth centered on the pelvis, the mean-per-joint-error (MPJPE)  focuses on joints extracted from the mesh, and the Procrustes-aligned MPJPE (PA-MPJPE) computes the error on joints after a rigid transformation that minimizes the error for each mesh. In order to evaluate the location estimate in the camera coordinates system we also compute an Abs-PVE (Absolute PVE) that is not centered on the pelvis, and the TA-PVE (Translation-Aligned PVE) that aligns the location of the scene and can only be computed when at least 2 humans are detected. On Hi4D and Harmony4D, we also evaluate the interaction with the Pair-PA-MPJPE that applies a single Procrustes-alignement to both humans altogether instead of doing it for each mesh separately, and compute the F1-score and Recall to evaluate the detection under occlusions. On the CMU-Panoptic dataset we compute the mean-root-positional-error (MRPE) to capture the accuracy of the translation, and report metrics normalized by the F1-score to account for detection. For evaluation on HMR tasks, following prior works, we match the predictions with the targets using an Hungarian matching based on the 2D joint locations. Metrics on MSCOCO are the keypoint average precisions (AP) and recalls (AR). 
To evaluate tracking, we report
the ID F1-score (IDF1) that evaluates how well the tracker maintains correct object identities over time, and the multiple object tracking accuracy (MOTA) that takes into account missed detections, false positives, and
identity switches. We use the evaluation code from~\cite{comotion}, that correctly accounts for the regions to ignore.

\subsection{Comparison with the state of the art}
\label{subsec:SOTA-comparison}

We evaluate \ours on in-the-wild HMR (Table~\ref{tab:3dpw-emdb}), interaction recovery (Table~\ref{tab:hi4d-harmony4d} and~\ref{tab:cmu_panoptic}), vertical FOV prediction (Table~\ref{tab:fov_comparison}), 2D alignment (Table~\ref{tab:coco_reprojection}) and tracking (Table~\ref{tab:tracking_results}) benchmarks.

\paragraph{In-the-wild HMR.} Table~\ref{tab:3dpw-emdb} compares \ours with one-shot multi-person HMR methods on the 3DPW test set and EMDB1, without any finetuning on the 3DPW training set. \ours significantly outperforms prior methods on both benchmarks, especially in terms of TA-PVE and Abs-PVE, which are the metrics that take into account the 3D localization in the camera space. AiOS particularly struggles to predict 3D locations because of its weak-perspective camera assumption. SAT-HMR also makes strong assumptions on the camera, but a fixed FOV of 60 degrees is much more realistic than a fixed focal length of 5000~mm. In particular, \ours outperforms Multi-HMR in terms of 3D localization on 3DPW while using a predicted camera.

\begin{table*}[t]
\centering
\caption{\textbf{Evaluation on CMU-Panoptic}.}
\vspace{-0.3cm}
\small
\renewcommand{\arraystretch}{1.0}
\begin{tabular}{l|cccc|cc}
\toprule
{Method} & {\footnotesize haggling} & {\footnotesize mafia} & {\footnotesize ultimatum} & {\footnotesize pizza} & \multicolumn{2}{c}{mean} \\
{\scriptsize $\dagger$ \textit{use official code+ckpt}} & \multicolumn{4}{c|}{\footnotesize N-MPJPE $\downarrow$} & {\footnotesize N-MPJPE $\downarrow$} & {\footnotesize N-MRPE $\downarrow$} \\
\midrule
3DCrowdNet~\cite{choi2022learning} & 115.4 & 143.1 & 136.6 & 142.7 & 134.0 & - \\
BEV~\cite{bev} & 93.5 & 106.9 & 116.6 & 129.1 & 112.9 & - \\
PSVT~\cite{psvt} & ~~91.4 & 100.9 & 118.8 & 124.8 & 109.0 & - \\
DPMesh~\cite{zhu2024dpmesh} & ~~97.2 & 109.8 & 114.3 & 120.5 & 110.4 & - \\
Scene~\cite{luvizon2023scene} & ~~84.5 & - & 108.9 & 133.2 & 108.9 & 217.1 \\
SAT-HMR$\dagger$~\cite{sathmr} & 107.4 & 134.2 & 126.9 & 126.8 & 123.8 & ~~83.7 \\
Multi-HMR$\dagger$~\cite{multihmr} & ~~\underline{73.1} & ~~\underline{85.8} & ~~\underline{94.0} & \textbf{~~84.6} & ~~\underline{84.4} & ~~\underline{58.1} \\
\textbf{\ours.a (ours)} & \textbf{~~68.9} & \textbf{~~81.3} & \textbf{~~82.7} & ~~\underline{88.9} & \textbf{~~81.5} & \textbf{~~26.9} \\
\bottomrule
\end{tabular}
\normalsize
\label{tab:cmu_panoptic}
\vspace{-0.2cm}
\end{table*}

\paragraph{Interactions.} We evaluate \ours in the context of close interactions in Table~\ref{tab:hi4d-harmony4d}. We achieve state-of-the-art results in pelvis-centered metrics while improving over the state of the art in interaction recovery and placement in the scene, even with a predicted camera. In the context of closely interacting humans, \ours achieves a F1-score of 100\% on Hi4D and detects 97\% of annotated humans in Harmony4D, making it the strongest detector together with~\cite{sathmr}. Table~\ref{tab:cmu_panoptic} shows that \ours with a predicted camera significantly outperforms prior works for multi-person detection and translation estimation on the CMU-Panoptic~\cite{joo2015panoptic} dataset.

\begin{table}[h]
\centering
\caption{\textbf{Comparison of predicted FOV} (in degrees).}
\vspace{-0.3cm}
\small
\renewcommand{\arraystretch}{1.0}
\begin{tabular}{l@{~}c@{~~}c}
\toprule
Method & 3DPW & EMDB \\
\midrule
Perspective fields~\cite{perspectivefields} & 14.0 & 12.8 \\
Ctrl-C~\cite{ctrlc} & ~~\underline{5.6} & ~~5.4 \\
WildCamera~\cite{wildcamera}  & ~~8.2 & ~~\textbf{2.3} \\
SPEC-camcalib~\cite{spec} & ~~8.8 & ~~5.9 \\
HumanFoV~\cite{camerahmr} & ~~\textbf{5.0} & ~~5.7 \\
\textbf{\ours.a (ours)} & ~~6.4 & ~~\underline{4.0} \\
\bottomrule
\end{tabular}
\normalsize
\label{tab:fov_comparison}
\vspace{-0.2cm}
\end{table}

\paragraph{Camera estimation.} Overall, \ours with predicted intrinsics outperforms Multi-HMR that uses the ground-truth camera, even on metrics in the camera space. This result is impressive, since our location predictions are obtained as an inverse projection using our predicted 2D location, depth, and focal length. To better understand this result, we evaluate the vertical FOV predictions on 3DPW and EMDB and compare it with FOV estimation methods in Table~\ref{tab:fov_comparison}. We achieve results comparable to specialized FOV estimation methods, which is why using a predicted camera has reasonable impact on our performance. In particular, while most methods are either accurate on 3DPW or EMDB, \ours.a is among the best approaches on both benchmarks. Aside from the 3D location, predicted human meshes are not conditioned on the camera. Pelvis-centered metrics (PVE, MPJPE, and PA-MPJPE) should then have the same value in both contexts, as centering on the pelvis eliminates location variations. While we observe this result on EMDB, we have slight difference in other datasets: this is due to the evaluation matching process relying on the reprojected joints.

\paragraph{Detection.} Table~\ref{tab:coco_reprojection} reports results on the MSCOCO val2017 set, using the MSCOCO API~\cite{coco}. Note that when not given camera intrinsic parameters, Multi-HMR adopts a fixed FOV of 60 degrees. 
\ours obtains the best re-projections, both in terms of precision and recall.
Again, note that \ours relies on a predicted camera while others use fixed or simplified camera models. Methods with learned perspective cameras face additional difficulty: the 3D pose must be consistent with varying cameras in order to reach high 2D precision. Despite this disadvantage, we demonstrate that jointly reasoning about detection, meshes, and camera leads to more reliable high-confidence predictions.

\begin{table}[h]
\centering
\caption{\textbf{Reprojected keypoints} on MSCOCO~\cite{coco}. 
$\ddagger$ denotes methods using datasets with 2D annotations for training.}
\vspace{-0.3cm}
\small
\resizebox{\linewidth}{!}{
\renewcommand{\arraystretch}{1.0}
\begin{tabular}{l|c@{~~}c@{~~}c@{~~}c@{~~}c@{~~}c}
\toprule
Method & {\footnotesize AP $\uparrow$} & {\footnotesize AP50 $\uparrow$} & {\footnotesize AP75 $\uparrow$} & {\footnotesize AR $\uparrow$} & {\footnotesize AR50 $\uparrow$} & {\footnotesize AR75 $\uparrow$} \\
\midrule
AiOS~\cite{aios} & \underline{44.9} & \underline{76.0} & \underline{46.4} & \underline{54.0} & \textbf{83.3} & \underline{57.7} \\
SAT-HMR$^\ddagger$~\cite{sathmr} & 42.1 & 73.8 & 43.8 & 49.5 & 78.4 & 53.0 \\
Multi-HMR~\cite{multihmr} & 31.2 & 60.0 & 29.1 & 39.0 & 66.0 & 39.8 \\
\textbf{\ours.b}$^\ddagger$ (ours) & \textbf{50.7} & \textbf{81.0} & \textbf{54.3} & \textbf{56.2} & \underline{83.2} & \textbf{60.9} \\
\bottomrule
\end{tabular}
}
\normalsize
\label{tab:coco_reprojection}
\vspace{-0.2cm}
\end{table}

\begin{table}
\centering
\caption{\textbf{Tracking results} on the PoseTrack21~\cite{posetrack21} validation set.}
\vspace{-0.3cm}
\small
\renewcommand{\arraystretch}{0.9}
\begin{tabular}{l@{~}c@{~~}c}
\toprule
Method & IDF1 $\uparrow$ & MOTA $\uparrow$ \\
\midrule
\multicolumn{3}{l}{\textit{Trained on HMR images}} \\
~~~~~~Humans4D~\cite{humans4d} & 72.5 & 60.2 \\
~~~~~~\textbf{\ours.b (ours)} & \textbf{75.2} & \textbf{65.5} \\[0.2cm]

\multicolumn{3}{l}{\textit{Trained on HMR videos}} \\
~~~~~~CoMotion~\cite{comotion} & \textbf{79.5} & \textbf{70.1} \\
\bottomrule
\end{tabular}
\normalsize
\label{tab:tracking_results}
\vspace{-0.2cm}
\end{table}

\begin{table*}[t]
\centering
\caption{\textbf{Ablation studies.} (Top left) \textbf{Ablation on the training data.} S stands for synthetic data, R for real images with pseudo-ground truth meshes, and 2D for datasets with 2D annotations. (Bottom left) \textbf{Ablation study of the matching strategy.} (Top right) \textbf{Square padding at training and inference time.}  (Bottom right) \textbf{Ablation on the predicted camera parameters.}}

\begin{minipage}[c]{0.48\linewidth}
\small

\begin{subtable}[c]{\linewidth}
\small
\centering
\resizebox{\linewidth}{!}{
\begin{tabular}{l|c@{~~}cc@{~~}cc@{~~}c}
\toprule
\multirow{3}{*}{Data} &
\multicolumn{2}{c}{3DPW} &
\multicolumn{2}{c}{Hi4D} &
\multicolumn{2}{c}{MSCOCO} \\
\cmidrule(lr){2-3} \cmidrule(lr){4-5} \cmidrule(lr){6-7}
 & {\scriptsize PVE $\downarrow$} & {\scriptsize PA-MPJPE $\downarrow$} & {\scriptsize MPJPE $\downarrow$} & {\scriptsize \makecell{Pair-PA-\\MPJPE $\downarrow$}} & {\scriptsize AP $\uparrow$} & {\scriptsize AR $\uparrow$} \\
\midrule
S & 93.6 & 48.2 & ~~\underline{64.4} & ~~\textbf{64.0} & 11.9 & 24.7 \\
R & \textbf{88.7} & \underline{45.7} & ~~67.9 & ~~75.7 & \underline{21.3} & 26.6 \\
S+2D & 94.1 & 68.5 & 107.0 & 109.5 & 12.4 & 26.2 \\
R+2D & 94.5 & 50.1 & 105.9 & 113.9 & 20.6 & \textbf{33.8} \\
S+R & \underline{90.5} & \textbf{45.6} & ~~\textbf{60.0} & ~~\underline{77.8} & \textbf{21.9} & 29.4 \\
S+R+2D & 91.6 & 46.9 & ~~76.5 & ~~84.2 & 16.8 & \underline{32.9} \\
\bottomrule
\end{tabular}
}
\end{subtable}

\vspace{0.3cm}

\begin{subtable}[t]{\linewidth}
\centering
\resizebox{\linewidth}{!}{
\begin{tabular}{l|c@{~~}cc@{~~}cc@{~~}c}
\toprule
\multirow{3}{*}{Matching} &
\multicolumn{2}{c}{3DPW} &
\multicolumn{2}{c}{Hi4D} &
\multicolumn{2}{c}{MSCOCO} \\
\cmidrule(lr){2-3} \cmidrule(lr){4-5} \cmidrule(lr){6-7}
 & {\scriptsize PVE $\downarrow$} & {\scriptsize PA-MPJPE $\downarrow$} & {\scriptsize MPJPE $\downarrow$} & {\scriptsize \makecell{Pair-PA-\\MPJPE $\downarrow$}} & {\scriptsize AP $\uparrow$} & {\scriptsize AR $\uparrow$} \\
\midrule
2D loc & ~~\textbf{90.5} & \textbf{45.6} & \textbf{60.0} & \underline{77.8} & \textbf{21.9} & \underline{29.4} \\
3D loc & ~~\underline{93.0} & \underline{46.8} & \underline{65.4} & \textbf{76.4} & \underline{17.0} & \textbf{30.4} \\
J3D & 104.5 & 55.1 & 83.1 & 84.8 & ~~8.6 & 16.7 \\
\bottomrule
\end{tabular}
}
\end{subtable}

\end{minipage}
\hfill
\begin{minipage}[c]{0.48\linewidth}
\small

\begin{subtable}[t]{\linewidth}
\centering
\resizebox{\linewidth}{!}{
\begin{tabular}{cc|c@{~~}cc@{~~}cc@{~~}c}
\toprule
\multirow{3}{*}{\makecell{Train \\padding}} & 
\multirow{3}{*}{\makecell{Test \\padding}} & 
\multicolumn{2}{c}{3DPW} &
\multicolumn{2}{c}{Hi4D} &
\multicolumn{2}{c}{MSCOCO} \\
\cmidrule(lr){3-4} \cmidrule(lr){5-6} \cmidrule(lr){7-8}
 &  & {\scriptsize PVE $\downarrow$} & {\scriptsize Abs-PVE $\downarrow$} & {\scriptsize Abs-PVE $\downarrow$} & {\scriptsize \makecell{Pair-PA-\\MPJPE $\downarrow$}} & {\scriptsize AP $\uparrow$} & {\scriptsize AR $\uparrow$} \\
\midrule
\textcolor{green!60!black}{\checkmark} & \textcolor{green!60!black}{\checkmark} & 90.5 & 45.6 & 60.0 & 77.8 & 21.9 & 29.4 \\
\textcolor{green!60!black}{\checkmark} & \textcolor{red!80!black}{$\times$} & 113.4 & 57.9 & 62.2 & 91.6 & 0.5 & 3.8 \\
\textcolor{red!80!black}{$\times$} & \textcolor{green!60!black}{\checkmark} & 89.7 & 46.3 & 61.7 & 77.8 & 19.8 & 32.2 \\
\textcolor{red!80!black}{$\times$} & \textcolor{red!80!black}{$\times$} & 85.8 & 43.8 & 61.5 & 74.7 & 17.7 & 30.9 \\
\bottomrule
\end{tabular}
}
\end{subtable}

\vspace{0.6cm}

\begin{subtable}[t]{\linewidth}
\centering
\resizebox{\linewidth}{!}{
\begin{tabular}{cc|c@{~~}cc@{~~}cc@{~~}c}
\toprule
\multirow{3}{*}{\makecell{Train \\ camera}} & 
\multirow{3}{*}{\makecell{Test \\ camera}} & 
\multicolumn{2}{c}{3DPW} &
\multicolumn{2}{c}{Hi4D} &
\multicolumn{2}{c}{MSCOCO} \\
\cmidrule(lr){3-4} \cmidrule(lr){5-6} \cmidrule(lr){7-8}
 & & {\scriptsize PVE $\downarrow$} & {\scriptsize Abs-PVE $\downarrow$} & {\scriptsize Abs-PVE $\downarrow$} & {\scriptsize \makecell{Pair-PA-\\MPJPE $\downarrow$}} & {\scriptsize AP $\uparrow$} & {\scriptsize AR $\uparrow$} \\
\midrule
GT & GT & 90.5 & \underline{243.7} & \textbf{148.0} & \underline{77.8} & -- & -- \\
GT & pred & 92.4 & 609.8 & 213.7 & 91.0 & \textbf{21.9} & \underline{29.4} \\
pred & pred & \textbf{86.2} & 577.3 & 220.0 & \textbf{72.7} & \underline{19.6} & \textbf{31.0} \\
pred & GT & \underline{89.0} & \textbf{243.4} & \underline{159.5} & 89.7 & -- & -- \\
\bottomrule
\end{tabular}
}
\end{subtable}

\end{minipage}

\label{tab:ablations}
\end{table*}

\paragraph{Tracking.} We report the IDF1 and MOTA metrics on the PoseTrack21 validation set in Table~\ref{tab:tracking_results}. \ours outperforms Humans4D, which also recovers meshes frame by frame and associate the detections over frames. In their case, the meshes are recovered with a two-stage framework (detection then single-person mesh estimation) and the association is done with an Hungarian matching on cost matrix combining the distances between predicted and estimated pelvis (as ours), the similarity between some handcrafted appearance features and finally the distances between some predicted poses and estimated pose. The latter pose prediction requires human motion sequences, in contrast to our approach that does not leverage any sequential information. We also report the performance of the recent CoMotion~\cite{comotion} approach that is trained to predict directly short HMR tracks on video clips, at the cost of a long stepwise training.

\subsection{Ablation study}\label{subsec:ablations}

We ablate the main components of \ours.
Ablation
experiments use images of size 672, a ViT-L backbone initialized with DINOv2~\cite{dinov2}, and the model is trained for 500k iterations with synthetic datasets and real images with pseudo-ground truth meshes. Additional ablations are in the supplementary material.

\paragraph{Training data.} We first investigate the role of training data. We denote \textbf{R} the real images with pseudo-ground truth meshes from~\cite{camerahmr}, \textbf{S} the synthetic datasets (AnnyOne and BEDLAM), and \textbf{2D} images from OpenImages~\cite{openimages} pseudo-labeled with ViTPose~\cite{vitpose} estimations. 
Results in Table~\ref{tab:ablations} show that on HMR benchmarks, \textbf{S} and \textbf{R} seem to benefit the most to mesh recovery overall, while using \textbf{2D} seems to degrade the 3D performance. This could be expected, as improvements in 2D and 3D performance often compete — improving one can lead to a decline in the other. When evaluating on MSCOCO, we realize that \textbf{S} is the most important to improve the precision, while \textbf{2D} improves significantly the recall by improving detections for individuals far away from the camera.

\paragraph{Matching strategy.} We test different matching strategies (Table~\ref{tab:ablations}) with a fixed training budget, relying either on 2D localization in pixels, 
on 3D location, or on 3D keypoints in the camera frame. While matching on 2D and 3D locations can be done without forwarding all queries to the parametric body model, 3D keypoints are more costly to obtain. Matching on 2D or 3D locations gives similar results, and matching on 3D joints gives much larger errors in all metrics. \ours uses 2D matching as it allows us to use training data without 3D annotations.

\paragraph{Varying aspect ratios.} We experiment using different aspect ratios at training, and evaluate the impact of square padding at inference, in Table~\ref{tab:ablations}. When training only with squared images, the performance decrease significantly when testing with diverse aspect ratios: this is because the model never saw this kind of images during training. While using squared images gives the best 2D performance, training and testing with diverse aspect ratios gives the best 3D metrics. We note that when training with diverse aspect ratios, testing with squared images has little impact on the performance.

\paragraph{Predicted camera.} We experiment using ground-truth or predicted camera intrinsic parameters at train and test times (Table~\ref{tab:ablations}). While the ground-truth camera is consistently better to estimate the location in the camera coordinates system, other metrics -- including the Pair-PA-MPJPE that measures interaction recoveries -- benefit from using the same settings during training and inference.
\section{Conclusion}
\label{sec:conclusion}

We have introduced \ours, a robust DETR-based multi-person HMR method leveraging the Anny parametric body model. 
It outputs strong detections, accurate localization with a predicted camera, precise human mesh recovery in scene- and pelvis-centered coordinate systems, and features for tracking.
Even if \ours can be trained on a single GPU in about a week, future works may explore more advanced DETR strategies~\cite{dabdetr, deim, groupdetr} for faster convergence. Other lines of improvement include modeling camera distortions~\cite{zolly, blade} to be more robust to images captured with extreme camera angles such as selfies.

{
    \small
    \bibliographystyle{ieeenat_fullname}
    \bibliography{main}

@String(PAMI = {IEEE Trans. Pattern Anal. Mach. Intell.})

@String(IJCV = {Int. J. Comput. Vis.})

@String(CVPR= {IEEE Conf. Comput. Vis. Pattern Recog.})

@String(ICCV= {Int. Conf. Comput. Vis.})

@String(ECCV= {Eur. Conf. Comput. Vis.})

@String(ICLR = {Int. Conf. Learn. Represent.})

@String(PAMI  = {IEEE TPAMI})

@String(IJCV  = {IJCV})

@String(CVPR  = {CVPR})

@String(ICCV  = {ICCV})

@String(ECCV  = {ECCV})

@String(ICLR  = {ICLR})

@inproceedings{hi4d,
      author = {Yin, Yifei and Guo, Chen and Kaufmann, Manuel and Zarate, Juan and Song, Jie and Hilliges, Otmar}, 
      title = {{Hi4D}: 4D Instance Segmentation of Close Human Interaction}, 
      booktitle = {CVPR},
      year = {2023}
      }

@InProceedings{3dpw,
author = {von Marcard, Timo and Henschel, Roberto and Black, Michael J. and Rosenhahn, Bodo and Pons-Moll, Gerard},
title = {Recovering Accurate 3D Human Pose in The Wild Using IMUs and a Moving Camera},
booktitle = {ECCV},
year = {2018}
}

@article{smpl,
      author = {Loper, Matthew and Mahmood, Naureen and Romero, Javier and Pons-Moll, Gerard and Black, Michael J.},
      title = {{SMPL}: A Skinned Multi-Person Linear Model},
      journal = {ACM Trans. Graphics},
      year = {2015}
    }

@inproceedings{bedlam,
  title={{BEDLAM}: A synthetic dataset of bodies exhibiting detailed lifelike animated motion},
  author={Black, Michael J and Patel, Priyanka and Tesch, Joachim and Yang, Jinlong},
  booktitle={CVPR},
  year={2023}
}

@inProceedings{hmr,
  title={End-to-end Recovery of Human Shape and Pose},
  author = {Angjoo Kanazawa
  and Michael J. Black
  and David W. Jacobs
  and Jitendra Malik},
  booktitle={CVPR},
  year={2018}
}

@inproceedings{humans4d,
  title={Humans in {4D}: Reconstructing and Tracking Humans with Transformers},
  author={Goel, Shubham and Pavlakos, Georgios and Rajasegaran, Jathushan and Kanazawa, Angjoo and Malik, Jitendra},
  booktitle={ICCV},
  year={2023}
}

@inproceedings{romp,
  title={Monocular, one-stage, regression of multiple 3d people},
  author={Sun, Yu and Bao, Qian and Liu, Wu and Fu, Yili and Black, Michael J and Mei, Tao},
  booktitle={ICCV},
  year={2021}
}

@inproceedings{multihmr,
    title={{Multi-HMR}: Multi-Person Whole-Body Human Mesh Recovery in a Single Shot},
    author={Baradel, Fabien and Armando, Matthieu and Galaaoui, Salma and Br{\'e}gier, Romain and Weinzaepfel, Philippe and Rogez, Gr{\'e}gory and Lucas, Thomas},
    booktitle={ECCV},
    year={2024}
}

@inproceedings{bev,
  title={Putting people in their place: Monocular regression of 3d people in depth},
  author={Sun, Yu and Liu, Wu and Bao, Qian and Fu, Yili and Mei, Tao and Black, Michael J},
  booktitle={CVPR},
  year={2022}
}

@inproceedings{centernet,
  title={Objects as Points},
  author={Zhou, Xingyi and Wang, Dequan and Kr{\"a}henb{\"u}hl, Philipp},
  booktitle={arXiv preprint arXiv:1904.07850},
  year={2019}
}

@article{dinov2,
  title={{DINOv2}: Learning Robust Visual Features without Supervision},
  author={Oquab, Maxime and Darcet, Timothée and Moutakanni, Theo and Vo, Huy V. and Szafraniec, Marc and Khalidov, Vasil and Fernandez, Pierre and Haziza, Daniel and Massa, Francisco and El-Nouby, Alaaeldin and Howes, Russell and Huang, Po-Yao and Xu, Hu and Sharma, Vasu and Li, Shang-Wen and Galuba, Wojciech and Rabbat, Mike and Assran, Mido and Ballas, Nicolas and Synnaeve, Gabriel and Misra, Ishan and Jegou, Herve and Mairal, Julien and Labatut, Patrick and Joulin, Armand and Bojanowski, Piotr},
  journal={TMLR},
  year={2023}
}

@inproceedings{vitpose,
  title={Vi{TP}ose: Simple Vision Transformer Baselines for Human Pose Estimation},
  author={Yufei Xu and Jing Zhang and Qiming Zhang and Dacheng Tao},
  booktitle={NeurIPS},
  year={2022},
}

@inproceedings{vit,
  title={An image is worth 16x16 words: Transformers for image recognition at scale},
  author={Dosovitskiy, Alexey and Beyer, Lucas and Kolesnikov, Alexander and Weissenborn, Dirk and Zhai, Xiaohua and Unterthiner, Thomas and Dehghani, Mostafa and Minderer, Matthias and Heigold, Georg and Gelly, Sylvain and others},
  booktitle={ICLR},
  year={2021}
}

@inproceedings{mpii,
  author = {Mykhaylo Andriluka and Leonid Pishchulin and Peter Gehler and Schiele, Bernt},
  title = {{2D} Human Pose Estimation: New Benchmark and State of the Art Analysis},
  booktitle = {CVPR},
  year = {2014},
}

@inproceedings{aios,
  title={{AiOS}: All-in-one-stage expressive human pose and shape estimation},
  author={Sun, Qingping and Wang, Yanjun and Zeng, Ailing and Yin, Wanqi and Wei, Chen and Wang, Wenjia and Mei, Haiyi and Leung, Chi-Sing and Liu, Ziwei and Yang, Lei and others},
  booktitle={CVPR},
  year={2024}
}

@inproceedings{camerahmr,
  title={{CameraHMR}: Aligning People with Perspective},
  author={Patel, Priyanka and Black, Michael J},
  booktitle={3DV},
  year={2025}
}

@inproceedings{detectNtrack,
  title={Detect-and-track: Efficient pose estimation in videos},
  author={Girdhar, Rohit and Gkioxari, Georgia and Torresani, Lorenzo and Paluri, Manohar and Tran, Du},
  booktitle={CVPR},
  year={2018}
}

@inproceedings{motionqueries,
    title={Motions as Queries: One-Stage Multi-Person Holistic Human Motion Capture},
    author={Liu, Kenkun and Fu, Yurong and Yuan, Weihao and Lin, Jing and Li, Peihao and Gu, Xiaodong and Qiu, Lingteng and Wang, Haoqian and Dong, Zilong and Han, Xiaoguang},
    booktitle={CVPR},
    year={2025}
}

@inproceedings{comotion,
  title={{CoMotion}: Concurrent Multi-person 3D Motion},
  author={Newell, Alejandro and Hu, Peiyun and Lipson, Lahav and Richter, Stephan R and Koltun, Vladlen},
  booktitle={ICLR},
  year={2025}
}

@article{sam2,
  title={{SAM} 2: Segment anything in images and videos},
  author={Ravi, Nikhila and Gabeur, Valentin and Hu, Yuan-Ting and Hu, Ronghang and Ryali, Chaitanya and Ma, Tengyu and Khedr, Haitham and R{\"a}dle, Roman and Rolland, Chloe and Gustafson, Laura and others},
  journal={arXiv preprint arXiv:2408.00714},
  year={2024}
}

@inproceedings{coco,
  title={{Microsoft COCO}: Common objects in context},
  author={Lin, Tsung-Yi and Maire, Michael and Belongie, Serge and Hays, James and Perona, Pietro and Ramanan, Deva and Doll{\'a}r, Piotr and Zitnick, C Lawrence},
  booktitle={ECCV},
  year={2014},
}

@inproceedings{phalp,
  title={Tracking people by predicting {3D} appearance, location and pose},
  author={Rajasegaran, Jathushan and Pavlakos, Georgios and Kanazawa, Angjoo and Malik, Jitendra},
  booktitle={CVPR},
  year={2022}
}

@inproceedings{detr,
  title={End-to-end object detection with transformers},
  author={Carion, Nicolas and Massa, Francisco and Synnaeve, Gabriel and Usunier, Nicolas and Kirillov, Alexander and Zagoruyko, Sergey},
  booktitle={ECCV},
  year={2020},
}

@article{anny,
  title={Human Mesh Modeling for {A}nny Body},
  author={Br{\'e}gier, Romain and Fiche, Gu{\'e}nol{\'e} and Bravo-S{\'a}nchez, Laura and Lucas, Thomas and Armando, Matthieu and Weinzaepfel, Philippe and Rogez, Gr{\'e}gory and Baradel, Fabien},
  journal={arXiv preprint arXiv:2511.03589},
  year={2025}
}

@inproceedings{transformer,
  title={Attention is all you need},
  author={Vaswani, Ashish and Shazeer, Noam and Parmar, Niki and Uszkoreit, Jakob and Jones, Llion and Gomez, Aidan N and Kaiser, {\L}ukasz and Polosukhin, Illia},
  booktitle={NeurIPS},
  year={2017}
}

@inproceedings{ghum,
  title={{GHUM} \& {GHUML}: Generative 3d human shape and articulated pose models},
  author={Xu, Hongyi and Bazavan, Eduard Gabriel and Zanfir, Andrei and Freeman, William T and Sukthankar, Rahul and Sminchisescu, Cristian},
  booktitle={CVPR},
  year={2020}
}

@inproceedings{smplx,
    title = {Expressive Body Capture: 3D Hands, Face, and Body from a Single Image},
    author = {Pavlakos, Georgios and Choutas, Vasileios and Ghorbani, Nima and Bolkart, Timo and Osman, Ahmed A. A. and Tzionas, Dimitrios and Black, Michael J.},
    booktitle = {CVPR},
    year = {2019}
}

@article{dinov3,
  title={{DINO}v3},
  author={Sim{\'e}oni, Oriane and Vo, Huy V and Seitzer, Maximilian and Baldassarre, Federico and Oquab, Maxime and Jose, Cijo and Khalidov, Vasil and Szafraniec, Marc and Yi, Seungeun and Ramamonjisoa, Micha{\"e}l and others},
  journal={TMLR},
  year={2026}
}

@article{aic,
  title={Can adversarial networks hallucinate occluded people with a plausible aspect?},
  author={Fulgeri, Federico and Fabbri, Matteo and Alletto, Stefano and Calderara, Simone and Cucchiara, Rita},
  journal={CVIU},
  year={2019},
}

@article{openimages,
  title={The open images dataset v4: Unified image classification, object detection, and visual relationship detection at scale},
  author={Kuznetsova, Alina and Rom, Hassan and Alldrin, Neil and Uijlings, Jasper and Krasin, Ivan and Pont-Tuset, Jordi and Kamali, Shahab and Popov, Stefan and Malloci, Matteo and Kolesnikov, Alexander and others},
  journal={IJCV},
  year={2020},
}

@misc{detectron2,
  author =       {Yuxin Wu and Alexander Kirillov and Francisco Massa and
                  Wan-Yen Lo and Ross Girshick},
  title =        {Detectron2},
  howpublished = {\url{https://github.com/facebookresearch/detectron2}},
  year =         {2019}
}

@inproceedings{sathmr,
  title={{SAT-HMR}: Real-Time Multi-Person 3D Mesh Estimation via Scale-Adaptive Tokens},
  author={Su, Chi and Ma, Xiaoxuan and Su, Jiajun and Wang, Yizhou},
  booktitle={CVPR},
  year={2025}
}

@inproceedings{emdb,
  title={{EMDB}: The electromagnetic database of global 3d human pose and shape in the wild},
  author={Kaufmann, Manuel and Song, Jie and Guo, Chen and Shen, Kaiyue and Jiang, Tianjian and Tang, Chengcheng and Z{\'a}rate, Juan Jos{\'e} and Hilliges, Otmar},
  booktitle={ICCV},
  year={2023}
}

@inproceedings{harmony4d,
  title={Harmony4d: A video dataset for in-the-wild close human interactions},
  author={Khirodkar, Rawal and Song, Jyun-Ting and Cao, Jinkun and Luo, Zhengyi and Kitani, Kris},
  booktitle={NeurIPS},
  year={2024}
}

@inproceedings{posetrack21,
  title={{PoseTrack21}: A dataset for person search, multi-object tracking and multi-person pose tracking},
  author={Doering, Andreas and Chen, Di and Zhang, Shanshan and Schiele, Bernt and Gall, Juergen},
  booktitle={CVPR},
  year={2022}
}

@inproceedings{maskrcnn,
  title={Mask {R}-{CNN}},
  author={He, Kaiming and Gkioxari, Georgia and Doll{\'a}r, Piotr and Girshick, Ross},
  booktitle={ICCV},
  year={2017}
}

@inproceedings{ssd,
  title={{SSD}: Single shot multibox detector},
  author={Liu, Wei and Anguelov, Dragomir and Erhan, Dumitru and Szegedy, Christian and Reed, Scott and Fu, Cheng-Yang and Berg, Alexander C},
  booktitle={ECCV},
  year={2016},
}

@inproceedings{yolo,
  title={{You Only Look Once}: Unified, real-time object detection},
  author={Redmon, Joseph and Divvala, Santosh and Girshick, Ross and Farhadi, Ali},
  booktitle={CVPR},
  year={2016}
}

@inproceedings{pymaf,
  title={{PyMAF}: 3d human pose and shape regression with pyramidal mesh alignment feedback loop},
  author={Zhang, Hongwen and Tian, Yating and Zhou, Xinchi and Ouyang, Wanli and Liu, Yebin and Wang, Limin and Sun, Zhenan},
  booktitle={ICCV},
  year={2021}
}

@inproceedings{smplerx,
    title={{SMPLer-X}: Scaling up expressive human pose and shape estimation},
    author={Cai, Zhongang and Yin, Wanqi and Zeng, Ailing and Wei, Chen and Sun, Qingping and Yanjun, Wang and Pang, Hui En and Mei, Haiyi and Zhang, Mingyuan and Zhang, Lei and Loy, Chen Change and Yang, Lei and Liu, Ziwei},
    booktitle={NeurIPS},
    year={2023}
}

@inproceedings{choi2022learning,
  title={Learning to estimate robust 3d human mesh from in-the-wild crowded scenes},
  author={Choi, Hongsuk and Moon, Gyeongsik and Park, JoonKyu and Lee, Kyoung Mu},
  booktitle={CVPR},
  year={2022}
}

@inproceedings{qiu2022dynamic,
  title={Dynamic graph reasoning for multi-person 3d pose estimation},
  author={Qiu, Zhongwei and Yang, Qiansheng and Wang, Jian and Fu, Dongmei},
  booktitle={ACM MM},
  year={2022}
}

@inproceedings{jiang2020coherent,
  title={Coherent reconstruction of multiple humans from a single image},
  author={Jiang, Wen and Kolotouros, Nikos and Pavlakos, Georgios and Zhou, Xiaowei and Daniilidis, Kostas},
  booktitle={CVPR},
  year={2020}
}

@inproceedings{zanfir2018monocular,
  title={Monocular 3d pose and shape estimation of multiple people in natural scenes-the importance of multiple scene constraints},
  author={Zanfir, Andrei and Marinoiu, Elisabeta and Sminchisescu, Cristian},
  booktitle={CVPR},
  year={2018}
}

@inproceedings{psvt,
  title={{PSVT}: End-to-end multi-person 3d pose and shape estimation with progressive video transformers},
  author={Qiu, Zhongwei and Yang, Qiansheng and Wang, Jian and Feng, Haocheng and Han, Junyu and Ding, Errui and Xu, Chang and Fu, Dongmei and Wang, Jingdong},
  booktitle={CVPR},
  year={2023}
}

@inproceedings{prompthmr,
  title={{PromptHMR}: Promptable Human Mesh Recovery},
  author={Wang, Yufu and Sun, Yu and Patel, Priyanka and Daniilidis, Kostas and Black, Michael J and Kocabas, Muhammed},
  booktitle={CVPR},
  year={2025}
}

@inproceedings{groupdetr,
  title={{Group DETR}: Fast detr training with group-wise one-to-many assignment},
  author={Chen, Qiang and Chen, Xiaokang and Wang, Jian and Zhang, Shan and Yao, Kun and Feng, Haocheng and Han, Junyu and Ding, Errui and Zeng, Gang and Wang, Jingdong},
  booktitle={ICCV},
  year={2023}
}

@inproceedings{deim,
  title={{DEIM}: Detr with improved matching for fast convergence},
  author={Huang, Shihua and Lu, Zhichao and Cun, Xiaodong and Yu, Yongjun and Zhou, Xiao and Shen, Xi},
  booktitle={CVPR},
  year={2025}
}

@inproceedings{dabdetr,
  title={{DAB}-{DETR}: Dynamic Anchor Boxes are Better Queries for {DETR}},
  author={Shilong Liu and Feng Li and Hao Zhang and Xiao Yang and Xianbiao Qi and Hang Su and Jun Zhu and Lei Zhang},
  booktitle={ICLR},
  year={2022},
}

@inproceedings{grouppose,
  title={Group pose: A simple baseline for end-to-end multi-person pose estimation},
  author={Liu, Huan and Chen, Qiang and Tan, Zichang and Liu, Jiang-Jiang and Wang, Jian and Su, Xiangbo and Li, Xiaolong and Yao, Kun and Han, Junyu and Ding, Errui and others},
  booktitle={ICCV},
  year={2023}
}

@inproceedings{shi2022end,
  title={End-to-end multi-person pose estimation with transformers},
  author={Shi, Dahu and Wei, Xing and Li, Liangqi and Ren, Ye and Tan, Wenming},
  booktitle={CVPR},
  year={2022}
}

@inproceedings{yang2023explicit,
  title={Explicit box detection unifies end-to-end multi-person pose estimation},
  author={Yang, Jie and Zeng, Ailing and Liu, Shilong and Li, Feng and Zhang, Ruimao and Zhang, Lei},
  booktitle={ICLR},
  year={2023}
}

@inproceedings{xiao2018simple,
  title={Simple baselines for human pose estimation and tracking},
  author={Xiao, Bin and Wu, Haiping and Wei, Yichen},
  booktitle={ECCV},
  year={2018}
}

@inproceedings{adam,
  author       = {Diederik P. Kingma and
                  Jimmy Ba},
  title        = {Adam: A Method for Stochastic Optimization},
  booktitle    = {ICLR},
  year         = {2015},
}

@article{ranftl2020towards,
  title={Towards robust monocular depth estimation: Mixing datasets for zero-shot cross-dataset transfer},
  author={Ranftl, Ren{\'e} and Lasinger, Katrin and Hafner, David and Schindler, Konrad and Koltun, Vladlen},
  journal={IEEE trans. PAMI},
  year={2020},
}

@article{mertan2022single,
  title={Single image depth estimation: An overview},
  author={Mertan, Alican and Duff, Damien Jade and Unal, Gozde},
  journal={Digital Signal Processing},
  year={2022},
}

@inproceedings{facil2019cam,
  title={{CAM-Convs}: Camera-aware multi-scale convolutions for single-view depth},
  author={Facil, Jose M and Ummenhofer, Benjamin and Zhou, Huizhong and Montesano, Luis and Brox, Thomas and Civera, Javier},
  booktitle={CVPR},
  year={2019}
}

@inproceedings{focalloss,
  title={Focal loss for dense object detection},
  author={Lin, Tsung-Yi and Goyal, Priya and Girshick, Ross and He, Kaiming and Doll{\'a}r, Piotr},
  booktitle={ICCV},
  year={2017}
}

@article{luo2019bagoftricks,
  title={A strong baseline and batch normalization neck for deep person re-identification},
  author={Luo, Hao and Jiang, Wei and Gu, Youzhi and Liu, Fuxu and Liao, Xingyu and Lai, Shenqi and Gu, Jianyang},
  journal={IEEE trans. Multimedia},
  year={2019},
}

@article{almazan2018re,
  title={Re-id done right: towards good practices for person re-identification},
  author={Almazan, Jon and Gajic, Bojana and Murray, Naila and Larlus, Diane},
  journal={arXiv preprint arXiv:1801.05339},
  year={2018}
}

@article{zheng2016person,
  title={Person re-identification: Past, present and future},
  author={Zheng, Liang and Yang, Yi and Hauptmann, Alexander G},
  journal={arXiv preprint arXiv:1610.02984},
  year={2016}
}

@inproceedings{zhong2018camera,
  title={Camera style adaptation for person re-identification},
  author={Zhong, Zhun and Zheng, Liang and Zheng, Zhedong and Li, Shaozi and Yang, Yi},
  booktitle={CVPR},
  year={2018}
}

@inproceedings{hiera,
  title={Hiera: A hierarchical vision transformer without the bells-and-whistles},
  author={Ryali, Chaitanya and Hu, Yuan-Ting and Bolya, Daniel and Wei, Chen and Fan, Haoqi and Huang, Po-Yao and Aggarwal, Vaibhav and Chowdhury, Arkabandhu and Poursaeed, Omid and Hoffman, Judy and others},
  booktitle={ICML},
  year={2023},
}

@inproceedings{hmar,
  title={Tracking people with 3D representations},
  author={Rajasegaran, Jathushan and Pavlakos, Georgios and Kanazawa, Angjoo and Malik, Jitendra},
  booktitle={NeurIPS},
  year={2021}
}

@inproceedings{spec,
  title={{SPEC}: Seeing people in the wild with an estimated camera},
  author={Kocabas, Muhammed and Huang, Chun-Hao P and Tesch, Joachim and M{\"u}ller, Lea and Hilliges, Otmar and Black, Michael J},
  booktitle={ICCV},
  year={2021}
}

@inproceedings{wildcamera,
  title={Tame a wild camera: In-the-wild monocular camera calibration},
  author={Zhu, Shengjie and Kumar, Abhinav and Hu, Masa and Liu, Xiaoming},
  booktitle={NeurIPS},
  year={2023}
}

@inproceedings{ctrlc,
  title={Ctrl-{C}: Camera calibration transformer with line-classification},
  author={Lee, Jinwoo and Go, Hyunsung and Lee, Hyunjoon and Cho, Sunghyun and Sung, Minhyuk and Kim, Junho},
  booktitle={ICCV},
  year={2021}
}

@inproceedings{perspectivefields,
  title={Perspective fields for single image camera calibration},
  author={Jin, Linyi and Zhang, Jianming and Hold-Geoffroy, Yannick and Wang, Oliver and Blackburn-Matzen, Kevin and Sticha, Matthew and Fouhey, David F},
  booktitle={CVPR},
  year={2023}
}

@inproceedings{6drot,
  title={On the continuity of rotation representations in neural networks},
  author={Zhou, Yi and Barnes, Connelly and Lu, Jingwan and Yang, Jimei and Li, Hao},
  booktitle={CVPR},
  year={2019}
}

@inproceedings{zolly,
  title={Zolly: Zoom focal length correctly for perspective-distorted human mesh reconstruction},
  author={Wang, Wenjia and Ge, Yongtao and Mei, Haiyi and Cai, Zhongang and Sun, Qingping and Wang, Yanjun and Shen, Chunhua and Yang, Lei and Komura, Taku},
  booktitle={ICCV},
  year={2023}
}

@inproceedings{blade,
  title={{BLADE}: Single-view Body Mesh Estimation through Accurate Depth Estimation},
  author={Wang, Shengze and Li, Jiefeng and Li, Tianye and Yuan, Ye and Fuchs, Henry and Nagano, Koki and De Mello, Shalini and Stengel, Michael},
  booktitle={CVPR},
  year={2025}
}

@inproceedings{joo2015panoptic,
  title={Panoptic studio: A massively multiview system for social motion capture},
  author={Joo, Hanbyul and Liu, Hao and Tan, Lei and Gui, Lin and Nabbe, Bart and Matthews, Iain and Kanade, Takeo and Nobuhara, Shohei and Sheikh, Yaser},
  booktitle={ICCV},
  year={2015}
}

@inproceedings{zhu2024dpmesh,
  title={Dpmesh: Exploiting diffusion prior for occluded human mesh recovery},
  author={Zhu, Yixuan and Li, Ao and Tang, Yansong and Zhao, Wenliang and Zhou, Jie and Lu, Jiwen},
  booktitle={CVPR},
  year={2024}
}

@inproceedings{luvizon2023scene,
  title={Scene-Aware 3D Multi-Human Motion Capture from a Single Camera},
  author={Luvizon, Diogo C and Habermann, Marc and Golyanik, Vladislav and Kortylewski, Adam and Theobalt, Christian},
  booktitle={Computer Graphics Forum},
  year={2023},
}
}

\clearpage
\appendix
\setcounter{section}{0}
\renewcommand{\thesection}{\Alph{section}}

\section{Additional ablations}
\label[appendix]{app:ablations}

\paragraph{Architecture.} We ablate the number of queries (Table~\ref{tab:query-count-results}) and of decoder blocks (Table~\ref{tab:dec-blocks-count-results}). We see that the impact of the number of queries is minimal, and keep 100 queries in \ours, similar to~\cite{sathmr}. While having a very small decoder with one or two blocks decreases the performance, the impact becomes less important from 4 blocks. \ours has 8 decoder blocks to have a good compromise between performance and efficiency.

\begin{table}[ht]
\centering
\caption{\textbf{Ablation study on the number of queries.}}
\vspace{-0.3cm}
\resizebox{\linewidth}{!}{
\begin{tabular}{cc@{~~}cc@{~~}cc@{~~}c}
\toprule
\multirow{2}{*}{\#queries} &
\multicolumn{2}{c}{3DPW} &
\multicolumn{2}{c}{Hi4D} &
\multicolumn{2}{c}{MSCOCO} \\
\cmidrule(lr){2-3} \cmidrule(lr){4-5} \cmidrule(lr){6-7}
 & \footnotesize{PVE $\downarrow$} & \footnotesize{PA-MPJPE $\downarrow$} & \footnotesize{MPJPE $\downarrow$} & \footnotesize{Pair-PA-MPJPE $\downarrow$} & \footnotesize{AP $\uparrow$} & \footnotesize{AR $\uparrow$}  \\
\midrule
50 & 90.8 & 45.2 & 62.6 & 74.4 & 21.5 & 34.4 \\
100 & 90.5 & 45.6 & 60.0 & 77.8 & 21.9 & 29.4 \\
200 & 90.0 & 45.0 & 66.4 & 76.6 & 23.3 & 34.5 \\
\bottomrule
\end{tabular}
}
\label{tab:query-count-results}
\end{table}

\begin{table}[ht]
\centering
\caption{\textbf{Ablation study on the number of blocks in the decoder.}}
\vspace{-0.3cm}
\resizebox{\linewidth}{!}{
\begin{tabular}{cc@{~~}cc@{~~}cc@{~~}c}
\toprule
\multirow{2}{*}{\makecell{\#blocks}} &
\multicolumn{2}{c}{3DPW} &
\multicolumn{2}{c}{Hi4D} &
\multicolumn{2}{c}{MSCOCO} \\
\cmidrule(lr){2-3} \cmidrule(lr){4-5} \cmidrule(lr){6-7}
 & \footnotesize{PVE $\downarrow$} & \footnotesize{PA-MPJPE $\downarrow$} & \footnotesize{MPJPE $\downarrow$} & \footnotesize{Pair-PA-MPJPE $\downarrow$} & \footnotesize{AP $\uparrow$} & \footnotesize{AR $\uparrow$} \\
\midrule
1 & 96.4 & 49 & 68.8 & 80.5 & 18.0 & 24.2 \\
2 & 91.1 & 47.0 & 68.4 & 81.6 & 20.4 & 30.5 \\
4 & 89.8 & 45.5 & 65.6 & 74.0 & 19.0 & 30.6 \\
8 & 90.5 & 45.6 & 60.0 & 77.8 & 21.9 & 29.4 \\
16 & 88.1 & 45.5 & 68.7 & 77.3 & 20.5 & 35.0 \\
\bottomrule
\end{tabular}
}
\vspace{-0.1cm}
\label{tab:dec-blocks-count-results}
\end{table}

\paragraph{2D angular loss.} Our proposed 2D angular loss improves the 3D performance, see Table~\ref{tab:angular-2d-loss}, while slightly degrading the reprojection. This result was expected because even if this is a reprojection loss, it is based on 3D ray directions. Aside from improving the 3D performance, it makes the convergence more robust, especially in the first steps of training where traditional 2D losses are usually discarded to avoid numerical instabilities.

\begin{table}[t]
\centering
\caption{\textbf{Ablation study of the angular 2D loss.}}
\vspace{-0.3cm}
\resizebox{\linewidth}{!}{
\begin{tabular}{cc@{~~}cc@{~~}cc@{~~}c}
\toprule
\multirow{2}{*}{\makecell{Angular \\ 2D loss}} &
\multicolumn{2}{c}{3DPW} &
\multicolumn{2}{c}{Hi4D} &
\multicolumn{2}{c}{MSCOCO} \\
\cmidrule(lr){2-3} \cmidrule(lr){4-5} \cmidrule(lr){6-7}
 & \footnotesize{PVE $\downarrow$} & \footnotesize{PA-MPJPE $\downarrow$} & \footnotesize{MPJPE $\downarrow$} & \footnotesize{Pair-PA-MPJPE $\downarrow$} & \footnotesize{AP $\uparrow$} & \footnotesize{AR $\uparrow$}  \\
\midrule
\textcolor{green!60!black}{\checkmark} & 90.5 & 45.6 & 60.0 & 77.8 & 21.9 & 29.4 \\
\textcolor{red!80!black}{$\times$} & 91.0 & 45.7 & 70.9 & 79.4 & 23.7 & 35.8 \\
\bottomrule
\end{tabular}
}
\vspace{-0.1cm}
\label{tab:angular-2d-loss}
\end{table}

\begin{figure*}[ht!]
\centering
\resizebox{0.8\linewidth}{!}{
\begin{tikzpicture}
\begin{groupplot}[
  group style={group size=2 by 3, vertical sep=25pt, horizontal sep=30pt},
  width=0.6\linewidth,
  height=0.4\linewidth,
  xtick={25,50,75},
  xmin=22, xmax=78,
  grid=both,
  grid style={line width=.1pt, draw=gray!20},
  major grid style={line width=.2pt, draw=gray!35},
  tick label style={font=\footnotesize},
  label style={font=\footnotesize},
  title style={font=\scriptsize, yshift=-6pt},
  every axis plot/.append style={line width=0.8pt, mark size=1.4pt, mark=*},
  cycle list={
    {blue, solid},
    {red, solid}
  }
]

\nextgroupplot[
  title={3DPW: PVE $\downarrow$},
  ytick={86,89,92}
]
\addplot coordinates {(25,90.7) (50,90.5) (75,87.0)};
\addplot coordinates {(25,93.0) (50,87.0) (75,85.5)};

\nextgroupplot[
  title={3DPW: PA-MPJPE $\downarrow$},
  ytick={44,47,50}
]
\addplot coordinates {(25,46.9) (50,45.6) (75,43.7)};
\addplot coordinates {(25,50.4) (50,45.1) (75,43.1)};

\nextgroupplot[
  title={Hi4D: MPJPE $\downarrow$},
  ytick={62,66,70}
]
\addplot coordinates {(25,69.5) (50,60.0) (75,65.4)};
\addplot coordinates {(25,71.4) (50,65.9) (75,62.8)};

\nextgroupplot[
  title={Hi4D: Pair-PA-MPJPE $\downarrow$},
  ytick={78,82,86}
]
\addplot coordinates {(25,86.8) (50,77.8) (75,79.6)};
\addplot coordinates {(25,78.7) (50,80.0) (75,75.1)};

\nextgroupplot[
  title={MSCOCO: AP $\uparrow$},
  ytick={18,21,24},
  xlabel={Epochs}
]
\addplot coordinates {(25,19.9) (50,21.9) (75,23.5)};
\addplot coordinates {(25,16.7) (50,21.9) (75,25.1)};

\nextgroupplot[
  title={MSCOCO: AR $\uparrow$},
  ytick={30,33,36},
  xlabel={Epochs}
]
\addplot coordinates {(25,29.5) (50,29.4) (75,31.2)};
\addplot coordinates {(25,31.3) (50,34.7) (75,36.9)};

\end{groupplot}
\end{tikzpicture}
}
\caption{
\textbf{Ablation on the backbone initialization.} We experiment using Dinov2-672 (\textcolor{blue}{blue}) and Dinov3-768 (\textcolor{red}{red}). 
}
\label{fig:dinov2-dinov3-results}
\end{figure*}
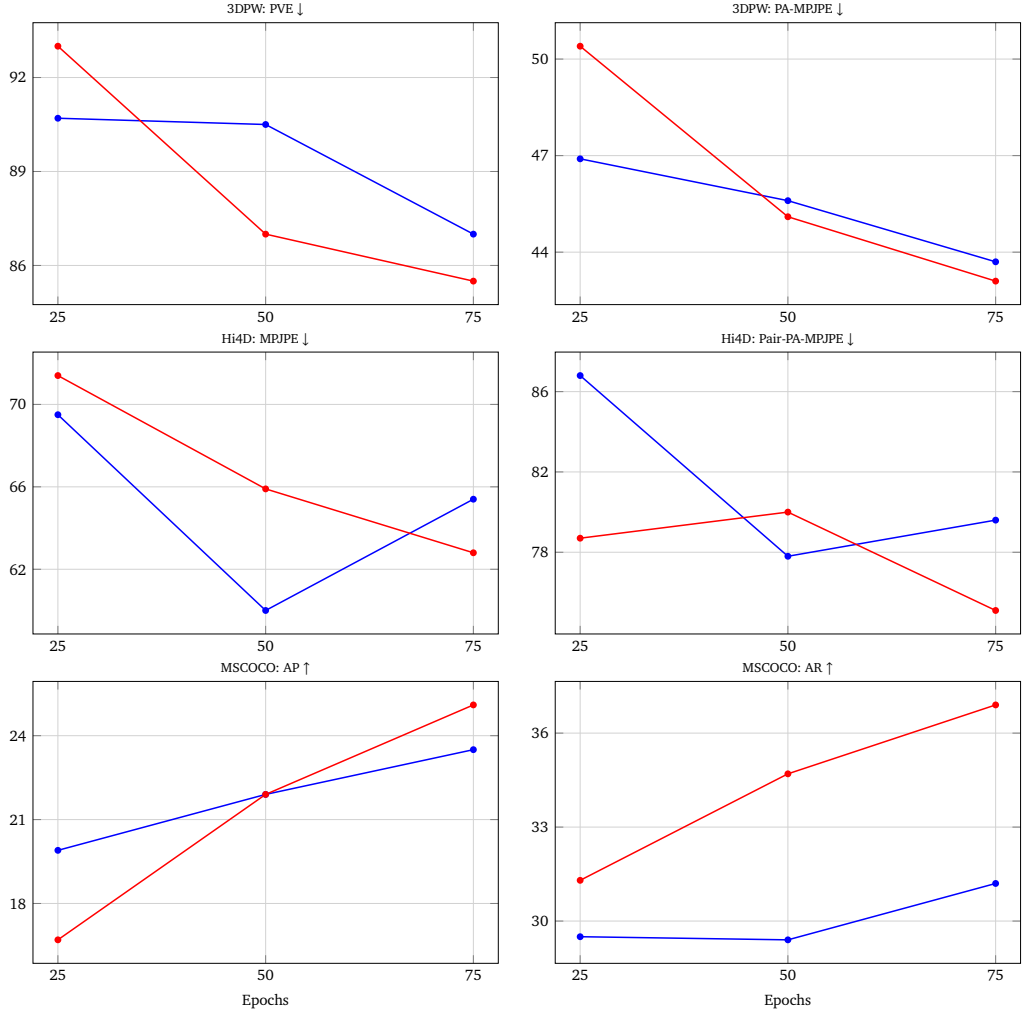

\paragraph{Backbone initialization.} We test initializing the backbone of our model with DINOv2~\cite{dinov2} and DINOv3~\cite{dinov3} in Figure~\ref{fig:dinov2-dinov3-results}. As DINOv3 uses patches of size 16 instead of 14, we increase the image dimension from 672 to 768 in order to have the same number of image tokens. While DINOv2 converges faster, and obtains better performance during the first iterations, DINOv3 then takes the lead and yields better performance on all benchmarks.

\section{Additional Comparisons}

We provide additional comparisons to the state of the art in \cref{tab:datasota}. First, we compare to BEV~\cite{bev} which was one of the first approach in predicting 3D location of people, and significantly outperforms it. Then we compare to Multi-HMR~\cite{multihmr} but trained with the Anny parametric body model~\cite{anny}, as reported in~\cite{anny}. This method uses exactly the same training data and parametric body model than ours.
We obtain overall consistent gains over Multi-HMR w. Anny on all datasets while being more robust in terms of detections and placements of people in the scene, see \cref{app:qualitative}.

\begin{table*}[t]
    \caption{\textbf{Additional comparisons} to the state of the art, with BEV and with Multi-HMR w. Anny. The latter method uses exactly the same training set and parametric body model than ours.}
    \label{tab:datasota}
    \vspace{-0.3cm}
    \centering
    \resizebox{\linewidth}{!}{
    \begin{tabular}{lccc|ccc|ccc}
    \hline
        & \multicolumn{3}{c|}{3DPW} & \multicolumn{3}{c|}{EMDB} & \multicolumn{3}{c}{Hi4D} \\[-0.1cm]
        & {\scriptsize PA-MPJPE $\downarrow$} & {\scriptsize MPJPE $\downarrow$} & {\scriptsize PVE $\downarrow$} & {\scriptsize PA-MPJPE $\downarrow$} & {\scriptsize MPJPE $\downarrow$} & {\scriptsize PVE $\downarrow$} & {\scriptsize PA-MPJPE $\downarrow$} & {\scriptsize MPJPE $\downarrow$} & {\scriptsize Pair-PA-MPJPE $\downarrow$} \\
    \hline
    BEV~\cite{bev} & 46.9 & 78.5 & 92.3 & 70.9 & 112.2 & 133.4 & 81.0 & 92.5 & 136.0 \\
    Multi-HMR w. Anny~\cite{multihmr,anny} & 41.8 & 71.5 & 83.2 & 48.5 & 71.5 & 83.4 & \textbf{48.7} & 66.6 & 80.0 \\
    \textbf{\ours (ours)}  & \textbf{40.2} &\textbf{68.2} & \textbf{80.1} & \textbf{48.3} & \textbf{68.5} & \textbf{78.7} & 49.1 & \textbf{61.7} & \textbf{79.5} \\
    \hline    
\end{tabular}
}
\end{table*}

\section{Additional Details}

\paragraph{Runtime.}
Without any particular optimization of the inference, \ours processes between 4 and 5 frames per second on a NVIDIA V100 GPU. Compiling the model with Pytorch significantly speeds-up inference: we achieve a frame rate of 20 on a NVIDIA V100 GPU, making \ours realtime despite using a ViT-Large backbone.

\paragraph{Ablation setup.}
To speed-up the experiments and decrease the computational cost, all ablations are performed in a restricted setting on a single V100 GPU for 500k iterations (around 4 days). While 3D metrics tend to converge quite fast and make little progress in later steps, the 2D re-projection keeps improving until the end of training, which explains the performance gap on MSCOCO between Table 4 and Table 5 of the main paper. 

When ablating the matching strategy, we again make the comparison with a similar computational budget, as the goal of this ablation is to show the efficiency of the matching predictions with the ground truth based solely on the translation. When using 3D joints obtained as output of the body model, we need to decrease the batch size and the model is trained for much less epochs as it is significantly slower. This explains the performance gap, we can reasonably think that matching on the 3D joints would lead to reasonable results when trained for much longer.

\paragraph{Evaluation.}  
Our method relies on the Anny parametric model~\cite{anny}. When evaluating on datasets with other body models, following the way SMPL-X-based methods evaluated on SMPL-based datasets, we use regressors between vertices. More precisely, we predict Anny meshes, that are converted to SMPL-X meshes using the regressor from~\cite{anny} between Anny vertices and SMPL-X vertices and we proceed to evaluation using SMPL-X meshes. Note that this is to our disadvantage as the regressor might introduce error (3.2mm cyclic error reported in~\cite{anny}) to our method when evaluating on these datasets.
When evaluating on SMPL-based datasets, we additionally use the SMPL-X to SMPL datasets.

\begin{figure*}[t]
    \centering
    \includegraphics[width=\textwidth]{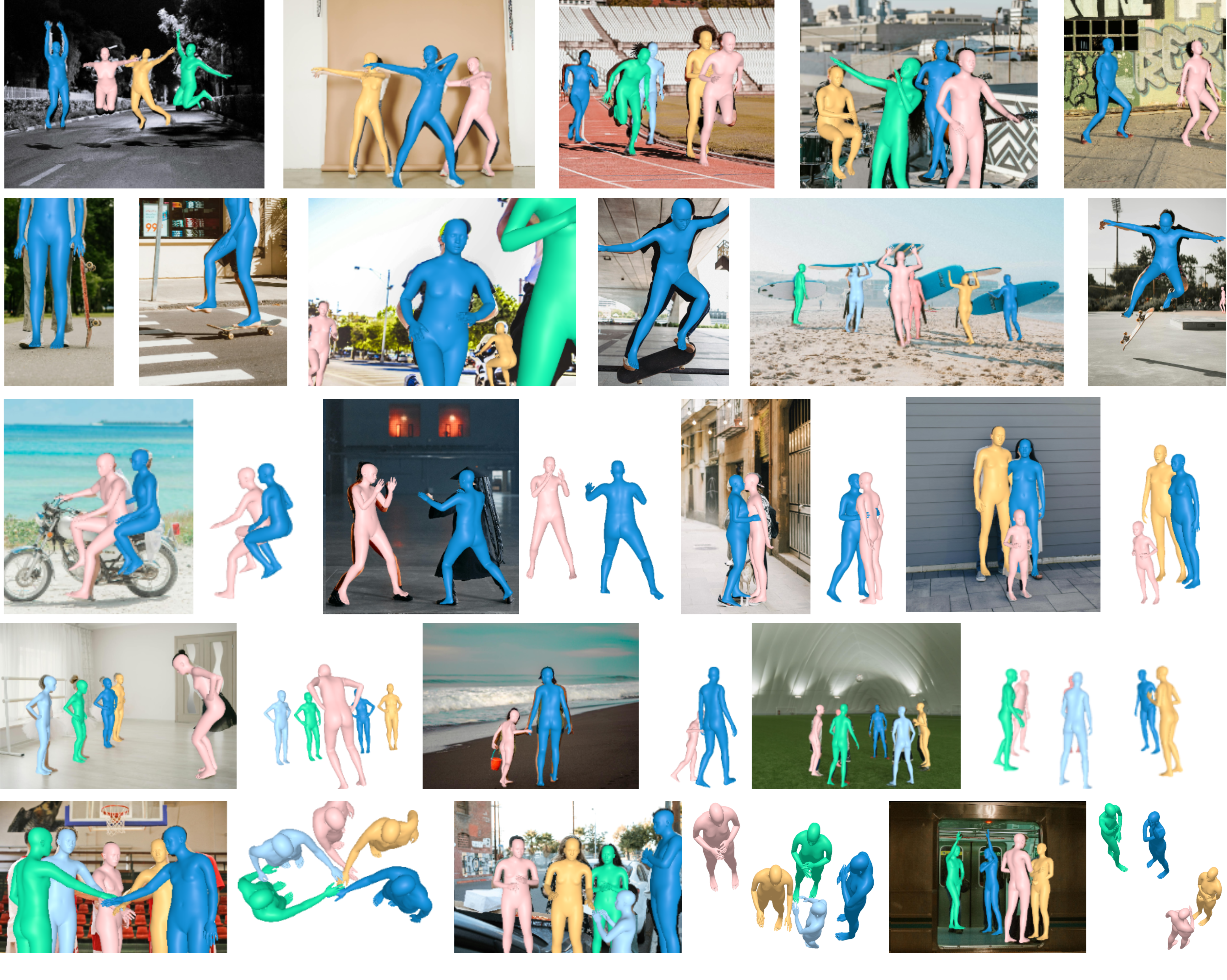} \\[-0.3cm]
    \caption{\textbf{Additional Qualitative Results.}}
    \vspace{-0.2cm}
    \label{fig:qualitative}
\end{figure*}

\section{Additional Qualitative Results}
\label{app:qualitative}

\paragraph{Qualitative results.} We visualize more results of \ours in Figure~\ref{fig:qualitative}. \ours achieves robust detections, with accurate translation estimation and re-projection, even in challenging scenarios.

\begin{figure*}[ht]
    \centering    \includegraphics[width=0.98\textwidth,height=0.9\textheight,keepaspectratio]{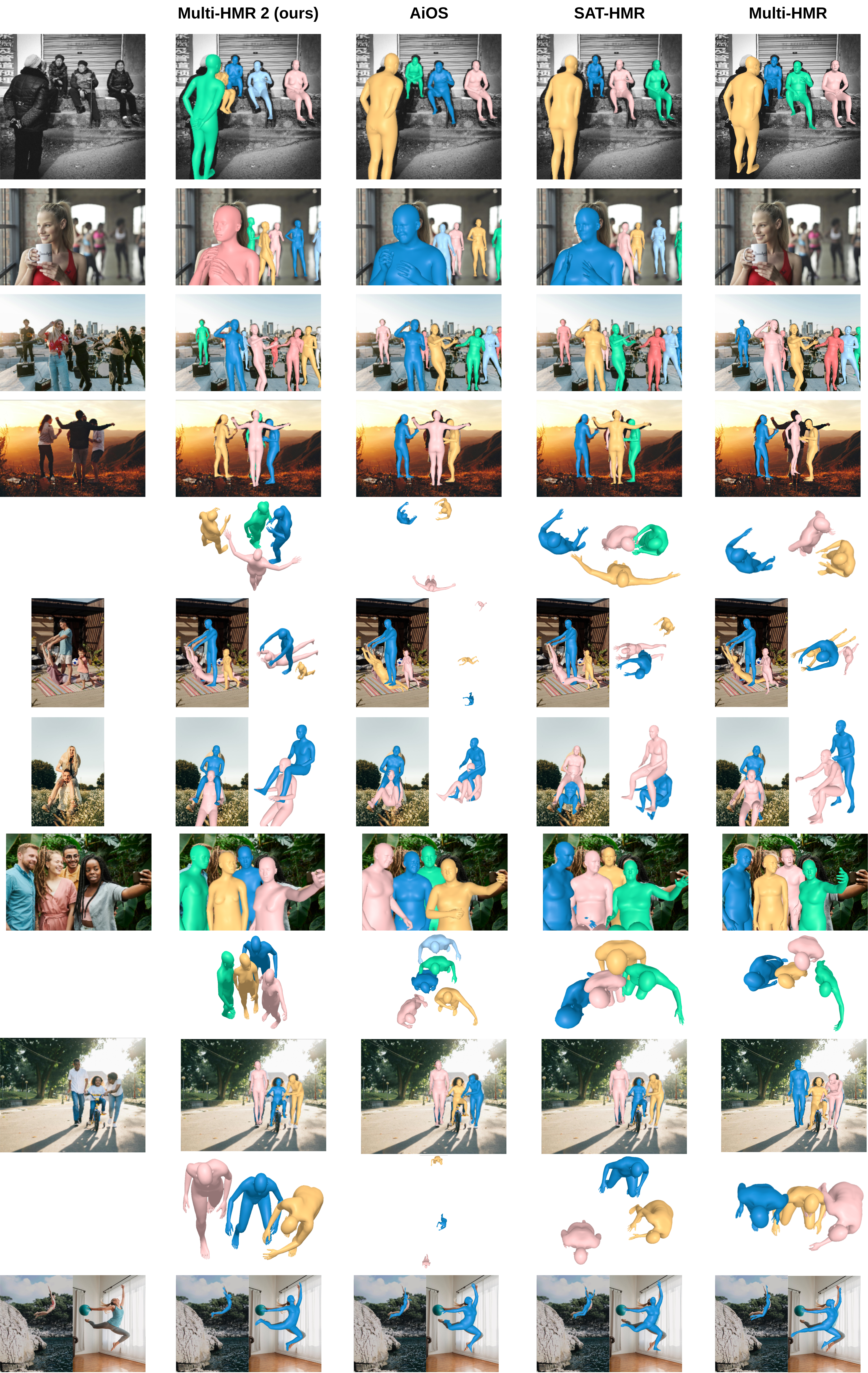}
    \caption{\textbf{Qualitative comparisons.}}
    \label{fig:qualitative_comparison}
\end{figure*}

\paragraph{Qualitative comparisons.} We compare \ours to other DETR-based multi-person HMR methods SAT-HMR~\cite{sathmr} and AiOS~\cite{aios} in Figure~\ref{fig:qualitative_comparison}. We see that \ours is particularly strong in terms of detection and 3D localization, especially in close interaction scenarios and with children, confirming the quantitative experiments of the main paper.

\end{document}